\documentclass[11pt,a4paper]{article}
\usepackage[hyperref]{acl2019}
\usepackage{times}
\usepackage{latexsym}
\usepackage{graphicx}
\usepackage{bm}
\usepackage{multirow}
\usepackage{booktabs}
\usepackage{siunitx}
\usepackage{fixmath}
\usepackage[skip=4pt]{caption}
\usepackage{tabularx}
\usepackage{tikz-dependency}
\usepackage{tikz}
\usepackage{amsfonts, amsmath, amsthm, amssymb} % For math fonts, symbols and environments

\usepackage{url}

\aclfinalcopy % Uncomment this line for the final submission

\renewcommand{\vec}[1]{\mathbold{#1}}
\newcommand{\mat}[1]{\mathbold{#1}}

\pgfkeys{%
/depgraph/reserved/edge style/.style = {% 
white, -, >=stealth, % arrow properties                                                                            
black, solid, line cap=round, % line properties
rounded corners=2, % make corners round
},%
}

\title{75 Languages, 1 Model: Parsing Universal Dependencies Universally}

\author{
    Dan Kondratyuk$^{1,2}$ \and
    Milan Straka$^{1}$ \\
    $^{1}$Charles University, Institute of Formal and Applied Linguistics\\
    $^{2}$Saarland University, Department of Computational Linguistics \\
    {\tt\footnotesize dankondratyuk@gmail.com, straka@ufal.mff.cuni.cz}
}

\date{}

\begin{document}
\maketitle
\begin{abstract}
    We present UDify, a multilingual multi-task model capable of accurately predicting universal part-of-speech, morphological features, lemmas, and dependency trees simultaneously for all 124 Universal Dependencies treebanks across 75 languages.
    By leveraging a multilingual BERT self-attention model pretrained on 104 languages, we found that fine-tuning it on all datasets concatenated together with simple softmax classifiers for each UD task can meet or exceed state-of-the-art UPOS, UFeats, Lemmas, (and especially) UAS, and LAS scores, without requiring any recurrent or language-specific components.
    We evaluate UDify for multilingual learning, showing that low-resource languages benefit the most from cross-linguistic annotations.
    We also evaluate for zero-shot learning, with results suggesting that multilingual training provides strong UD predictions even for languages that neither UDify nor BERT have ever been trained on.
    % Finally, we provide evidence to explain why pretrained self-attention networks might excel in multilingual dependency parsing.
    Code for UDify is available at \url{https://github.com/hyperparticle/udify}.
\end{abstract}

\section{Introduction}

\begin{figure}[!ht]
    \centering
    \includegraphics[width=\linewidth]{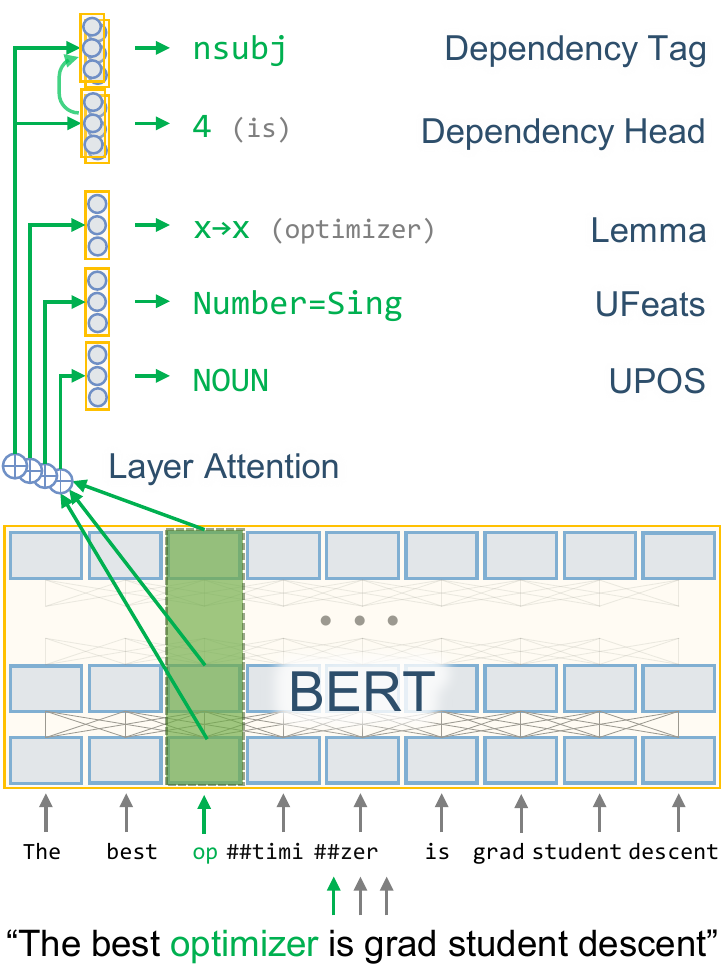}
    \caption{\label{fig:udify-architecture}
        An illustration of the UDify network architecture with task-specific layer attention, inputting word tokens and outputting UD annotations for each token.
    }
\end{figure}

In the absence of annotated data for a given language, it can be considerably difficult to create models that can parse the language's text accurately.
Multilingual modeling presents an attractive way to circumvent this low-resource limitation.
In a similar way learning a new language can enhance the proficiency of a speaker's previous languages \cite{abu2010advantages}, a model which has access to multilingual information can begin to learn generalizations across languages that would not have been possible through monolingual data alone.
Works such as \newcite{mcdonald2011multi, naseem2012selective, duong2015low, ammar2016many, de2018parameter, kitaev2018multilingual, mulcaire2019polyglot} consistently demonstrate how pairing the training data of similar languages can boost evaluation scores of models predicting syntactic information like part-of-speech and dependency trees.
Multilinguality not only can improve a model's evaluation performance, but can also reduce the cost of training multiple models for a collection of languages \cite{johnson2017google, smith201882}.

However, scaling to a higher number of languages can often be problematic.
Without an ample supply of training data for the considered languages, it can be difficult to form appropriate generalizations and especially difficult if those languages are distant from each other.
But recent techniques in language model pretraining can profit from a drastically larger supply of unsupervised text, demonstrating the capability of transferring contextual sentence-level knowledge to boost the parsing accuracy of existing NLP models \cite{howard2018universal, Peters:2018, devlin2018bert}.

One such model, BERT \cite{devlin2018bert}, introduces a self-attention (Transformer) network that results in state-of-the-art parsing performance when fine-tuning its contextual embeddings.
And with the release of a multilingual version pretrained on the entirety of the top 104 resourced languages of Wikipedia, BERT is remarkably capable of capturing an enormous collection of cross-lingual syntactic information.
Conveniently, these languages nearly completely overlap with languages supported by the Universal Dependencies treebanks, which we will use to demonstrate the ability to scale syntactic parsing up to 75 languages and beyond.

The Universal Dependencies (UD) framework provides syntactic annotations consistent across a large collection of languages \cite{11234/1-2895, zeman-EtAl:2018:K18-2}.
This makes it an excellent candidate for analyzing syntactic knowledge transfer across multiple languages.
UD offers tokenized sentences with annotations ideal for multi-task learning, including lemmas ({\sc Lemmas}), treebank-specific part-of-speech tags ({\sc XPOS}), universal part-of-speech tags ({\sc UPOS}), morphological features ({\sc UFeats}), and dependency edges and labels ({\sc Deps}) for each sentence.

We propose UDify, a semi-supervised multi-task self-attention model automatically producing UD annotations in any of the supported UD languages.
To accomplish this, we perform the following:

\begin{enumerate}
    \item We input all sentences into a pretrained multilingual BERT network to produce contextual embeddings, introduce task-specific layer-wise attention similar to ELMo \cite{Peters:2018}, and decode each UD task simultaneously using softmax classifiers.
    \item We apply a heavy amount of regularization to BERT, including input masking, increased dropout, weight freezing, discriminative fine-tuning, and layer dropout.
    \item We train and fine-tune the model on the entirety of UD by concatenating {\it all} available training sets together.
\end{enumerate}

We evaluate our model with respect to UDPipe Future, one of the winners of the CoNLL 2018 Shared Task on Multilingual Parsing from Raw Text to Universal Dependencies \cite{straka:2018:K18-2, zeman-EtAl:2018:K18-2}.
In addition, we analyze languages that multilingual training benefits prediction the most, and evaluate the model for zero-shot learning, i.e., treebanks which do not have a training set.
Finally, we provide evidence from our experiments and other related work to help explain why pretrained self-attention networks excel in multilingual dependency parsing.

Our work uses the AllenNLP library built for the PyTorch framework.
Code for UDify and a release of the fine-tuned BERT weights are available at \url{https://github.com/hyperparticle/udify}.

\section{Multilingual Multi-Task Learning}

In this section, we detail the multilingual training setup and the UDify multi-task model architecture. See Figure~\ref{fig:udify-architecture} for an architecture diagram.

\subsection{Multilingual Pretraining with BERT}

We leverage the provided BERT base multilingual cased pretrained model\footnote{\url{https://github.com/google-research/bert/blob/master/multilingual.md}}, with a self-attention network of 12 layers, 12 attention heads per layer, and hidden dimensions of 768 \cite{devlin2018bert}.
The model was trained by predicting randomly masked input words on the entirety of the top 104 languages with the largest Wikipedias.
BERT uses a wordpiece tokenizer \cite{wu2016google}, which segments all text into (unnormalized) sub-word units.

\subsection{Cross-Linguistic Training Issues}

Table~\ref{table:vocab-size} displays a list of vocabulary sizes, indicating that UD treebanks possess nearly 1.6M unique tokens combined.
To sidestep the problem of a ballooning vocabulary, we use BERT's wordpiece tokenizer directly for all inputs.
UD expects predictions to be along word boundaries, so we take the simple approach of applying the tokenizer to each word using UD's provided segmentation.
For prediction, we use the outputs of BERT corresponding to the first wordpiece per word, ignoring the rest\footnote{We found last, max, or average pooling of the wordpieces were not any better or worse for evaluation. \newcite{kitaev2018multilingual} report similar results.}.

In addition, the XPOS annotations are not universal across languages, or even across treebanks.
Because each treebank can possess a different annotation scheme for XPOS which can slow down inference, we omit training and evaluation of XPOS from our experiments.

\begin{table}[tbp]
    \small
    \begin{center}
    \begin{tabular}{@{}lr@{}}
    \toprule
        \sc Token & \sc Vocab Size \\
    \midrule
        Word Form & 1,588,655 \\
        BERT Wordpieces & 119,547 \\
        UPOS & 17 \\
        XPOS & 19,126 \\
        UFeats & 23,974 \\
        Lemmas (tags) & 109,639 \\
        Deps & 251 \\
    \bottomrule
    \end{tabular}
    \end{center}
    \caption{\label{table:vocab-size} 
        Vocabulary sizes of words and tags over all of UD v2.3, with a total of 12,032,309 word tokens and 668,939 sentences.
    }
\end{table}

\subsection{Multi-Task Learning with UD} \label{sec:multi-task-learning-ud}

For predicting UD annotations, we employ a multi-task network based on UDPipe Future \cite{straka:2018:K18-2}, but with all embedding, encoder, and projection layers replaced with BERT.
The remaining components include the prediction layers for each task detailed below, and layer attention (see Section~\ref{sec:layer-attention}).
Then we compute softmax cross entropy loss on the output logits to train the network.
For more details on reasons behind architecture choices, see Appendix~\ref{sec:appendix}.

{\bf UPOS} \, As is standard for neural sequence tagging, we apply a softmax layer along each word input, computing a probability distribution over the tag vocabulary to predict the annotation string.

{\bf UFeats} \, Identical to UPOS prediction, we treat each UFeats string as a separate token in the vocabulary.
We found this to produce higher evaluation accuracy than predicting each morphological feature separately.
Only a small subset of the full Cartesian product of morphological features is valid, eliminating invalid combinations.\\

{\bf Lemmas} \, Similar to \newcite{chrupala2006simple, muller2015joint}, we reduce the problem of lemmatization to a sequence tagging problem by predicting a class representing an edit script, i.e., the sequence of character operations to transform the word form to the lemma.
To precompute the tags, we first find the longest common substring between the form and the lemma, and then compute the shortest edit script converting the prefix and suffix of the form into the prefix and suffix of the lemma using the Wagner--Fischer algorithm \cite{wagner1974string}.
Upon predicting a lemma edit script, we apply the edit operations to the word form to produce the final lemma.
See also \newcite{straka:2018:K18-2} for more details.
We chose this approach over a sequence-to-sequence architecture like \newcite{bergmanis2018context} or \newcite{kondratyuk2018lemmatag}, as this significantly reduces training efficiency.

{\bf Deps} \, We use the graph-based biaffine attention parser developed by \newcite{dozat2016deep, dozat2017stanford}, replacing the bidirectional LSTM layers with BERT.
The final embeddings are projected through arc-head and arc-dep feedforward layers, which are combined using biaffine attention to produce a probability distribution of arc heads for each word.
We then decode each tree with the Chu-Liu/Edmonds algorithm \cite{chu1965shortest, edmonds1967optimum}.

\section{Fine-Tuning BERT on UD Annotations}

We employ several strategies for fine-tuning BERT for UD prediction, finding that regularization is absolutely crucial for producing a high-scoring network.

\subsection{Layer Attention} \label{sec:layer-attention}

Empirical results suggest that when fine-tuning BERT, combining the output of the last several layers is more beneficial for the downstream tasks than just using the last layer \cite{devlin2018bert}.
Instead of restricting the model to any subset of layers, we devise a simple layer-wise dot-product attention where the network computes a weighted sum of all intermediate outputs of the 12 BERT layers using the same weights for each token.
This is similar to how ELMo mixes the output of multiple recurrent layers \cite{Peters:2018}.

More formally, let $\vec{w}_i$ be a trainable scalar for BERT embeddings $\mat{\operatorname{\bf BERT}}_{ij}$ at layer $i$ with a token at position $j$, and let $c$ be a trainable scalar.
We compute contextual embeddings $\vec{e}^{\operatorname{(task)}}$ such that

\begin{equation} \label{eq:layer-attention}
    \vec{e}_{j}^{\operatorname{(task)}} = c \sum_i \mat{\operatorname{\bf BERT}}_{ij} \cdot \operatorname{softmax}(\vec{w})_i
\end{equation}

To prevent the UD classifiers from overfitting to the information in any single layer, we devise {\bf layer dropout}, where at each training step, we set each parameter $\vec{w}_i$ to $-\infty$ with probability $0.1$.
This effectively redistributes probability mass to all other layers, forcing the network to incorporate the information content of all BERT layers.
We compute layer attention per task, using one set of $c, \vec{w}$ parameters for each of UPOS, UFeats, Lemmas, and Deps.

\subsection{Transfer Learning with ULMFiT} \label{sec:ulmfit}

The ULMFiT strategy defines several useful methods for fine-tuning a network on a pretrained language model \cite{howard2018universal}. We apply the same methods, with a few minor modifications.

We split the network into two parameter groups, i.e., the parameters of BERT and all other parameters.
We apply discriminative fine-tuning, setting the base learning rate of BERT to be $5e^{-5}$ and $1e^{-3}$ everywhere else.
We also freeze the BERT parameters for the first epoch to increase training stability.

While ULMFiT recommends decaying the learning rate linearly after a linear warmup, we found that this is prone to training divergence in self-attention networks, introducing vanishing gradients and underfitting.
Instead, we apply an inverse square root learning rate decay with linear warmup (Noam) seen in training Transformer networks for machine translation \cite{vaswani2017attention}.

\subsection{Input Masking}

The authors of BERT recommend not to mask words randomly with {\tt [MASK]} when fine-tuning the network.
However, we discovered that masking often reduces the tendency of the classifiers to overfit to BERT by forcing the network to rely on the context of surrounding words.
This {\it word dropout} strategy has been observed in other works showing improved test performance on a variety of NLP tasks \cite{iyyer2015deep, bowman2016generating, clark2018semi, straka:2018:K18-2}.

\section{Experiments}

We evaluate UDify with respect to every test set in each treebank.
As there are too many results to fit within one page, we display a salient subset of scores and compare them with UDPipe Future.
The full results are listed in Appendix~\ref{sec:appendix}.

We do not directly reference metrics from other models in the CoNLL 2018 Shared Task, as the tables of results do not assume gold word segmentation and may not provide a fair comparison.
Instead, we retrained the open source UDPipe Future model using gold segmentation and report results here due to its architectural similarity to UDify and its strong performance.

Note that the UDPipe Future baseline does not itself use BERT. Evaluation of BERT utilization in UDPipe Future can be found in \citet{udpipe_bert}.

\subsection{Datasets}

For all experiments, we use the full Universal Dependencies v2.3 corpus available on LINDAT \cite{11234/1-2895}.
We omit the evaluation of datasets that do not have their training annotations freely available, i.e., Arabic NYUAD (ar\_nyuad), English ESL (en\_esl), French FTB (fr\_ftb), Hindi English HEINCS (qhe\_heincs), and Japanese BCCWJ (ja\_bccwj).

To train the multilingual model, we concatenate all available training sets together, similar to \newcite{mcdonald2011multi}.
Before each epoch, we shuffle all sentences and feed mixed batches of sentences to the network, where each batch may contain sentences from any language or treebank, for a total of 80 epochs\footnote{
We train on a GTX 1080 Ti for approximately 25 days. See Appendix \ref{sec:udify-hyperparameters} for more details}.

\subsection{Hyperparameters}

A summary of hyperparameters can be found in Table~\ref{table:hyperparameters} in Appendix~\ref{sec:udify-hyperparameters}.

\begin{table}[!t]
    \fontsize{8}{10}\selectfont
    \begin{center}
    \setlength{\tabcolsep}{2pt}
    \begin{tabularx}{\linewidth}{@{}Xlrrrrr@{}}
    \toprule
        \sc Treebank & \sc Model & \sc UPOS & \sc Feats & \sc Lem & \sc UAS & \sc LAS \\
    \midrule
    \multirow{4}{*}[\normalbaselineskip]{\shortstack[l]{Czech PDT \\ (cs\_pdt)}}
        & UDPipe     &      99.18 &      97.23 &  \bf 99.02 &      93.33 &      91.31 \\
    \addlinespace[2pt]
        & Lang       &      99.18 &      96.87 &      98.72 &      94.35 &      92.41 \\
        & UDify      &      99.18 &      96.85 &      98.56 &      94.73 &      92.88 \\
        & UDify+Lang &  \bf 99.24 &  \bf 97.44 &      98.93 &  \bf 95.07 &  \bf 93.38 \\
    \addlinespace[5pt]
    \multirow{4}{*}[\normalbaselineskip]{\shortstack[l]{German GSD \\ (de\_gsd)}}
        & UDPipe     &      94.48 &      90.68 &  \bf 96.80 &      85.53 &      81.07 \\
        & Lang       &      94.77 &      91.73 &      96.34 &      87.54 &      83.39 \\
        & UDify      &      94.55 &      90.65 &      94.82 &      87.81 &      83.59 \\
        & UDify+Lang &  \bf 95.29 &  \bf 91.94 &      96.74 &  \bf 88.11 &  \bf 84.13 \\
    \addlinespace[5pt]
    \multirow{4}{*}[\normalbaselineskip]{\shortstack[l]{English EWT \\ (en\_ewt )}}
        & UDPipe     &      96.29 &      97.10 &  \bf 98.25 &      89.63 &      86.97 \\
    \addlinespace[2pt]
        & Lang       &  \bf 96.82 &  \bf 97.27 &      97.97 &  \bf 91.70 &  \bf 89.38 \\
        & UDify      &      96.21 &      96.17 &      97.35 &      90.96 &      88.50 \\
        & UDify+Lang &      96.57 &      96.96 &      97.90 &      91.55 &      89.06 \\
    \addlinespace[5pt]
    \multirow{4}{*}[\normalbaselineskip]{\shortstack[l]{Spanish AnCora \\ (es\_ancora)}}
        & UDPipe     &  \bf 98.91 &  \bf 98.49 &  \bf 99.17 &      92.34 &      90.26 \\
    \addlinespace[2pt]
        & Lang       &      98.60 &      98.14 &      98.52 &      92.82 &      90.52 \\
        & UDify      &      98.53 &      97.84 &      98.09 &      92.99 &      90.50 \\
        & UDify+Lang &      98.68 &      98.25 &      98.68 &  \bf 93.35 &  \bf 91.28 \\
    \addlinespace[5pt]
    \multirow{4}{*}[\normalbaselineskip]{\shortstack[l]{French GSD \\ (fr\_gsd)}}
        & UDPipe     &      97.63 &  \bf 97.13 &  \bf 98.35 &      90.65 &      88.06 \\
    \addlinespace[2pt]
        & Lang       &  \bf 98.05 &      96.26 &      97.96 &      92.77 &      90.61 \\
        & UDify      &      97.83 &      96.59 &      97.48 &  \bf 93.60 &  \bf 91.45 \\
        & UDify+Lang &      97.96 &      96.73 &      98.17 &      93.56 &      91.45 \\
    \addlinespace[5pt]
    \multirow{4}{*}[\normalbaselineskip]{\vspace{-10pt} \shortstack[l]{Russian \\ SynTagRus \\ (ru\_syntagrus)}}
        & UDPipe     &  \bf 99.12 &  \bf 97.57 &  \bf 98.53 &      93.80 &      92.32 \\
    \addlinespace[2pt]
        & Lang       &      98.90 &      96.58 &      95.16 &      94.40 &      92.72 \\
        & UDify      &      98.97 &      96.35 &      94.43 &      94.83 &      93.13 \\
        & UDify+Lang &      99.08 &      97.22 &      96.58 &  \bf 95.13 &  \bf 93.70 \\
    \midrule
    \multirow{4}{*}[\normalbaselineskip]{\shortstack[l]{Belarusian HSE \\ (be\_hse)}}
        & UDPipe     &      93.63 &      73.30 &      87.34 &      78.58 &      72.72 \\
    \addlinespace[2pt]
        & Lang       &      95.88 &      76.12 &      84.52 &      83.94 &      79.02 \\
        & UDify      &  \bf 97.54 &  \bf 89.36 &      85.46 &  \bf 91.82 &  \bf 87.19 \\
        & UDify+Lang &      97.25 &      85.02 &  \bf 88.71 &      90.67 &      86.98 \\
    \addlinespace[5pt]
    \multirow{4}{*}[\normalbaselineskip]{\shortstack[l]{Buryat BDT \\ (bxr\_bdt)}}
        & UDPipe     &      40.34 &      32.40 &      58.17 &      32.60 &      18.83 \\
    \addlinespace[2pt]
        & Lang       &      52.54 &      37.03 &      54.64 &      29.63 &      15.82 \\
        & UDify      &  \bf 61.73 &  \bf 47.86 &  \bf 61.06 &  \bf 48.43 &  \bf 26.28 \\
        & UDify+Lang &      61.73 &      42.79 &      58.20 &      33.06 &      18.65 \\
    \addlinespace[5pt]
    \multirow{4}{*}[\normalbaselineskip]{\vspace{-10pt} \shortstack[l]{Upper Sorbian \\ UFAL \\ (hsb\_ufal)}}
        & UDPipe     &      62.93 &      41.10 &      68.68 &      45.58 &      34.54 \\
    \addlinespace[2pt]
        & Lang       &      73.70 &      46.28 &      58.02 &      39.02 &      28.70 \\
        & UDify      &      84.87 &      48.63 &  \bf 72.73 &  \bf 71.55 &  \bf 62.82 \\
        & UDify+Lang &  \bf 87.58 &  \bf 53.19 &      71.88 &      71.40 &      60.65 \\
    \addlinespace[5pt]
    \multirow{4}{*}[\normalbaselineskip]{\shortstack[l]{Kazakh KTB \\ (kk\_ktb)}}
        & UDPipe     &      55.84 &      40.40 &      63.96 &      53.30 &      33.38 \\
    \addlinespace[2pt]
        & Lang       &      73.52 &      46.60 &      57.84 &      50.38 &      32.61 \\
        & UDify      &  \bf 85.59 &  \bf 65.14 &  \bf 77.40 &  \bf 74.77 &  \bf 63.66 \\
        & UDify+Lang &      81.32 &      60.50 &      67.30 &      69.16 &      53.14 \\
    \addlinespace[5pt]
    \multirow{4}{*}[\normalbaselineskip]{\shortstack[l]{Lithuanian HSE \\ (lt\_hse)}}
        & UDPipe     &      81.70 &      60.47 &  \bf 76.89 &      51.98 &      42.17 \\
    \addlinespace[2pt]
        & Lang       &      83.40 &      54.34 &      58.77 &      51.23 &      38.96 \\
        & UDify      &  \bf 90.47 &  \bf 68.96 &      67.83 &  \bf 79.06 &  \bf 69.34 \\
        & UDify+Lang &      84.53 &      56.98 &      58.21 &      58.40 &      39.91 \\
    \bottomrule
    \end{tabularx}
    \end{center}
    \caption{\label{table:main-results} 
        Test set scores for a subset of high-resource (top) and low-resource (bottom) languages in comparison to UDPipe Future without BERT, with 3 UDify configurations:
        {\bf Lang}, fine-tune on the treebank.
        {\bf UDify}, fine-tune on all UD treebanks combined.
        {\bf UDify+Lang}, fine-tune on the treebank using BERT weights saved from fine-tuning on all UD treebanks combined.
    }
\end{table}

\begin{table}[!ht]
    \fontsize{8}{10}\selectfont
    \begin{center}
    \setlength{\tabcolsep}{3pt}
    \begin{tabularx}{\linewidth}{@{}lXrrrrr@{}}
    \toprule
        \sc Model & \sc Configuration & \sc UPOS & \sc Feats & \sc Lem & \sc UAS & \sc LAS \\
    \midrule
    UDPipe & w/o BERT          & \bf 93.76 & \bf 91.04 & \bf 94.63 &     84.37 &     79.76 \\
    \addlinespace
    UDify  & Task Layer Attn   &     93.40 &     88.72 &     90.41 & \bf 85.69 & \bf 80.43 \\
    UDify  & Global Layer Attn &     93.12 &     87.53 &     89.03 &     85.07 &     79.49 \\
    UDify  & Sum Layers        &     93.02 &     87.20 &     88.70 &     84.97 &     79.33 \\
    \bottomrule
    \end{tabularx}
    \end{center}
    \caption{\label{table:results-ablation}
        Ablation comparing the average of scores over all treebanks: task-specific layer attention (4 sets of $c,\vec{w}$ computed for the 4 UD tasks), global layer attention (one set of $c,\vec{w}$ for all tasks), and simple sum of layers ($c = 1$ and $\vec{w} = \vec{1}$).
    }
\end{table}

\begin{table}[!t]
    \fontsize{8}{10}\selectfont
    \begin{center}
    \setlength{\tabcolsep}{3pt}
    \begin{tabularx}{\linewidth}{@{}Xrrrrr@{}}
    \toprule
         \sc Treebank & \sc UPOS & \sc Feats & \sc Lem & \sc UAS & \sc LAS \\
    \midrule
         \bf Breton KEB \hfill\hfill br\_keb & 63.67 &  46.75 &  53.15 & 63.97 & 40.19 \\
         \bf Tagalog TRG \hfill\hfill tl\_trg & 61.64 &  35.27 &  75.00 & 64.73 & 39.38 \\
         Faroese OFT \hfill\hfill fo\_oft & 77.86 &  35.71 &  53.82 & 69.28 & 61.03 \\
         Naija NSC \hfill\hfill pcm\_nsc & 56.59 &  52.75 &  97.52 & 47.13 & 33.43 \\
         Sanskrit UFAL \hfill\hfill sa\_ufal & 40.21 &  18.45 &  37.60 & 41.73 & 19.80 \\
    \bottomrule
    % \toprule
    %      \sc Treebank & Model & \sc UPOS & \sc Feats & \sc Lem & \sc UAS & \sc LAS \\
    % \midrule
    % \multirow{4}{*}[\normalbaselineskip]{\shortstack[l]{\bf Breton KEB \\ \bf (br\_keb)}}
    %         & UDify & 63.67 &  46.75 &  53.15 & 63.97 & 40.19 \\
    % \addlinespace[2pt]
    % \multirow{4}{*}[\normalbaselineskip]{\shortstack[l]{\bf Tagalog TRG \\ (tl\_trg)}}
    %       & UDify & 61.64 &  35.27 &  75.00 & 64.73 & 39.38 \\
    % \addlinespace[2pt]
    % \multirow{4}{*}[\normalbaselineskip]{\shortstack[l]{ \\ }}
    %      Faroese OFT (fo\_oft)     & UDify & 77.86 &  35.71 &  53.82 & 69.28 & 61.03 \\
    % \addlinespace[2pt]
    % \multirow{4}{*}[\normalbaselineskip]{\shortstack[l]{ \\ }}
    %      Naija NSC (pcm\_nsc)      & UDify & 56.59 &  52.75 &  97.52 & 47.13 & 33.43 \\
    % \addlinespace[2pt]
    % \multirow{4}{*}[\normalbaselineskip]{\shortstack[l]{ \\ }}
    %      Sanskrit UFAL (sa\_ufal)  & UDify & 40.21 &  18.45 &  37.60 & 41.73 & 19.80 \\
    % \bottomrule
    \end{tabularx}
    \end{center}
    \caption{\label{table:zero-shot-results} 
        Test set results for zero-shot learning, i.e., no UD training annotations available. 
        Languages that are pretrained with BERT are bolded.
    }
\end{table}

\begin{table}[!ht]
    \fontsize{8}{10}\selectfont
    \begin{center}
    \begin{tabular}{@{}llr@{}}
    \toprule
         \sc Treebank & \sc Model & \sc UUAS \\
    \midrule
    % \multirow{4}{*}[\normalbaselineskip]{\shortstack[l]{English EWT \\ (en\_ewt)}}
          English EWT (en\_ewt) & BERT  & 65.48 \\
                                & BERT+finetune en\_ewt & \bf 79.67 \\
    \bottomrule
    \end{tabular}
    \end{center}
    \caption{\label{table:probe-comparison}
        UUAS test scores calculated on the predictions produced by the syntactic structural probe \cite{hewitt2019structural} using the English EWT treebank, on the unmodified multilingual cased BERT model and the same BERT model fine-tuned on the treebank.
    }
\end{table}

\subsection{Probing for Syntax}

\newcite{hewitt2019structural} introduce a structural probe for identifying dependency structures in contextualized word embeddings.
This probe evaluates whether syntax trees (i.e., unlabeled undirected dependency trees) can be easily extracted as a global property of the embedding space using a linear transformation of the network's contextual word embeddings.
The probe trains a weighted adjacency matrix on the layers of contextual embeddings produced by BERT, identifying a linear transformation where squared L2 distance between embedding vectors encodes the distance between words in the parse tree.
Edges are decoded by computing the minimum spanning tree on the weight matrix (the lowest sum of edge distances).

We train the structural probe on unmodified and fine-tuned BERT using the default hyperparameters of \newcite{hewitt2019structural} to evaluate whether the representations affected by fine-tuning BERT on dependency trees would more closely match the structure of these trees.

\section{Results}

We show scores of {\sc UPOS}, UFeats ({\sc Feats}), and Lemma ({\sc Lem}) accuracies, along with unlabeled and labeled attachment scores ({\sc UAS}, {\sc LAS}) evaluated using the offical CoNLL 2018 Shared Task evaluation script.\footnote{\url{https://universaldependencies.org/conll18/evaluation.html}}
Results for a salient subset of high-resource and low-resource languages are shown in Table~\ref{table:main-results}, with a comparison between UDPipe Future and UDify fine-tuning on all languages. In addition, the table compares UDify with fine-tuning on either a single language or both languages (fine-tuning multilingually, then fine-tuning on the language with the saved multilingual weights) to provide a reference point for multilingual influences on UDify.
We provide a full table of scores for all treebanks in Appendix~\ref{sec:full-results}.

\begin{figure}[tbp]
    \centering
    \includegraphics[width=\linewidth]{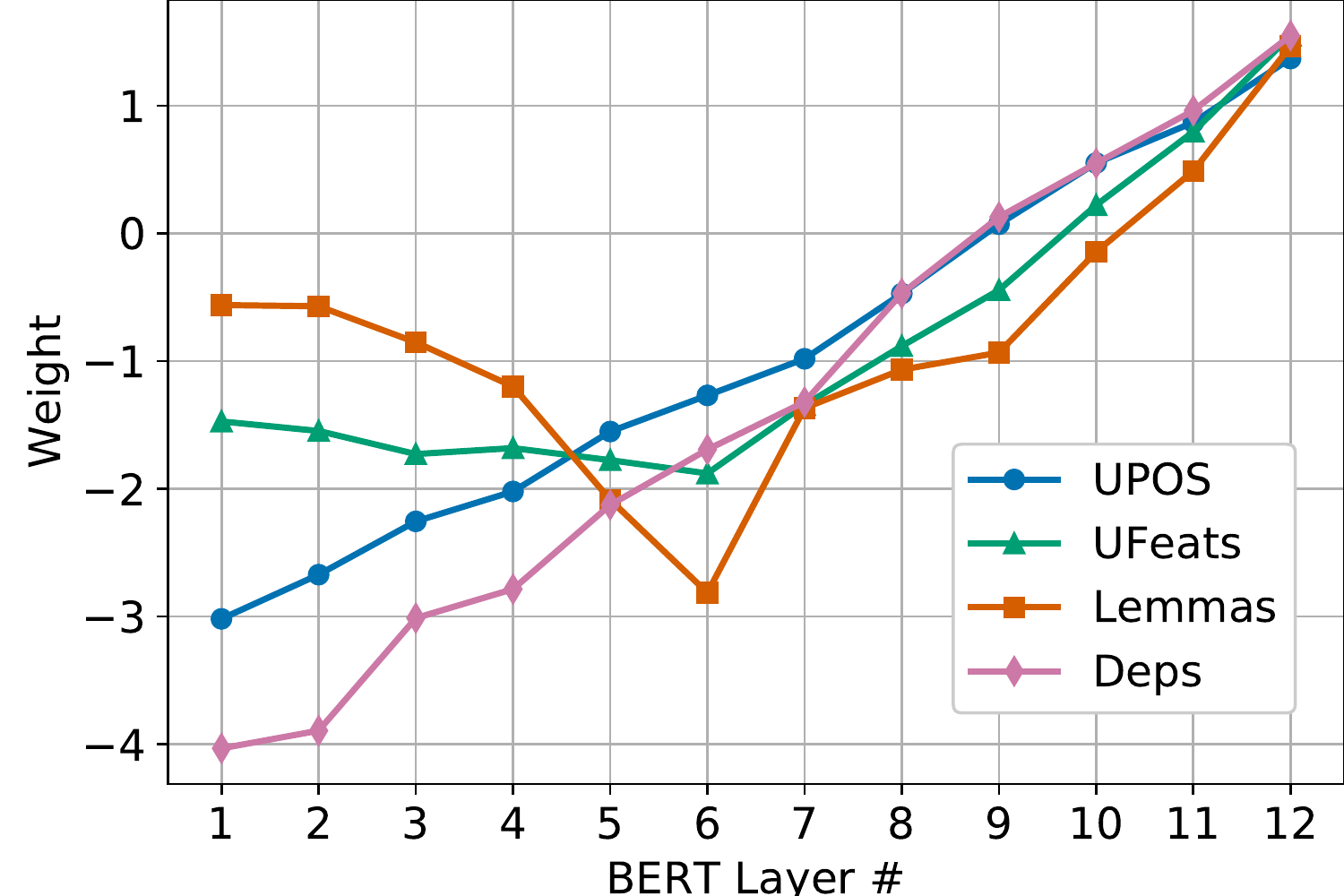}
    \caption{\label{fig:bert-layer-weights}
        The unnormalized BERT layer attention weights $\vec{w}_i$ contributing to layer $i$ for each task after training.
        A linear change in weight scales each BERT layer exponentially due to the softmax in Equation~\ref{eq:layer-attention}
    }
\end{figure}

A more comprehensive overview is shown in Table~\ref{table:results-ablation}, comparing different attention strategies applied to UDify.
We display an average of scores over all (89) treebanks with a training set.
For zero-shot learning evaluation, Table~\ref{table:zero-shot-results} displays a subset of test set evaluations of treebanks that do not have a training set, i.e., Breton, Tagalog, Faroese, Naija, and Sanskrit.
We plot the layer attention weights $\vec{w}$ after fine-tuning BERT in Figure~\ref{fig:bert-layer-weights}, showing a set of weights per task.
And Table~\ref{table:probe-comparison} compares the unlabeled undirected attachment scores (UUAS) of dependency trees produced using a structural probe on both the unmodified multilingual cased BERT model and the extracted BERT model fine-tuned on the English EWT treebank.

\section{Discussion}

In this section, we discuss the most notable features of the results.

\subsection{Model Performance}

On average, UDify reveals a strong set of results that are comparable in performance with the state-of-the-art in parsing UD annotations.
UDify excels in dependency parsing, exceeding UDPipe Future by a large margin especially for low-resource languages.
UDify slightly underperforms with respect to Lemmas and Universal Features, likely due to UDPipe Future additionally using character-level embeddings \cite{santos2014learning, ling2015finding, ballesteros2015improved, kim2016character}, while (for simplicity) UDify does not.
Additionally, UDify severely underperforms the baseline on a few low-resource languages, e.g., cop\_scriptorum.
We surmise that this is due to using mixed batches on an unbalanced training set, which skews the model towards predicting larger treebanks more accurately.
However, we find that fine-tuning on the treebank individually with BERT weights saved from UDify eliminates most of these gaps in performance.

\begin{figure*}[htbp]
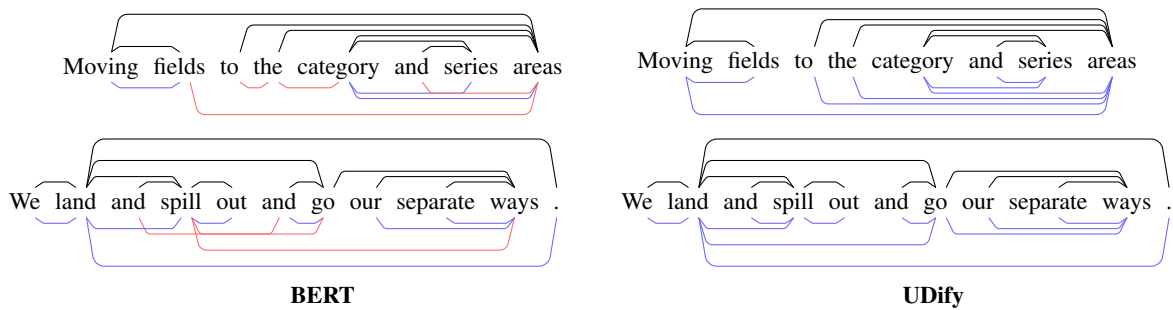

    \centering
    \small
    \begin{minipage}[b]{0.4\textwidth}
        \begin{dependency}[hide label, edge unit distance=.5ex]
          \begin{deptext}[column sep=0.05cm]
          Moving\& fields\& to\& the\& category\& and\& series\& areas \\
        \end{deptext}
        \depedge{1}{2}{.}
        \depedge{1}{8}{.}
        \depedge{3}{8}{.}
        \depedge{4}{8}{.}
        \depedge{5}{7}{.}
        \depedge{5}{8}{.}
        \depedge{6}{7}{.}
        \depedge[edge style={blue!60!}, edge below]{5}{7}{.}
        \depedge[edge style={blue!60!}, edge below]{1}{2}{.}
        \depedge[edge style={blue!60!}, edge below]{5}{8}{.}
        \depedge[edge style={red!60!}, edge below]{4}{5}{.}
        \depedge[edge style={red!60!}, edge below]{2}{8}{.}
        \depedge[edge style={red!60!}, edge below]{3}{4}{.}
        \depedge[edge style={red!60!}, edge below]{6}{8}{.}
        \end{dependency}
    \end{minipage}
    \hspace{1cm}
    \begin{minipage}[b]{0.4\textwidth}
        \begin{dependency}[hide label, edge unit distance=.5ex]
          \begin{deptext}[column sep=0.05cm]
          Moving\& fields\& to\& the\& category\& and\& series\& areas \\
        \end{deptext}
        \depedge{1}{2}{.}
        \depedge{1}{8}{.}
        \depedge{3}{8}{.}
        \depedge{4}{8}{.}
        \depedge{5}{7}{.}
        \depedge{5}{8}{.}
        \depedge{6}{7}{.}
        \depedge[edge style={blue!60!}, edge below]{5}{8}{.}
        \depedge[edge style={blue!60!}, edge below]{3}{8}{.}
        \depedge[edge style={blue!60!}, edge below]{5}{7}{.}
        \depedge[edge style={blue!60!}, edge below]{4}{8}{.}
        \depedge[edge style={blue!60!}, edge below]{1}{8}{.}
        \depedge[edge style={blue!60!}, edge below]{1}{2}{.}
        \depedge[edge style={blue!60!}, edge below]{6}{7}{.}
        \end{dependency}
    \end{minipage}
    
    \hspace{-1cm}
    \begin{minipage}[b]{0.4\textwidth}
        \begin{dependency}[hide label, edge unit distance=.5ex]
          \begin{deptext}[column sep=0.05cm]
          We\& land\& and\& spill\& out\& and\& go\& our\& separate\& ways\& . \\
        \end{deptext}
        \depedge{1}{2}{.}
        \depedge{2}{4}{.}
        \depedge{2}{7}{.}
        \depedge{2}{11}{.}
        \depedge{3}{4}{.}
        \depedge{4}{5}{.}
        \depedge{6}{7}{.}
        \depedge{7}{10}{.}
        \depedge{8}{10}{.}
        \depedge{9}{10}{.}
        \depedge[edge style={red!60!}, edge below]{3}{6}{.}
        \depedge[edge style={blue!60!}, edge below]{2}{4}{.}
        \depedge[edge style={blue!60!}, edge below]{6}{7}{.}
        \depedge[edge style={red!60!}, edge below]{4}{7}{.}
        \depedge[edge style={blue!60!}, edge below]{1}{2}{.}
        \depedge[edge style={blue!60!}, edge below]{9}{10}{.}
        \depedge[edge style={blue!60!}, edge below]{4}{5}{.}
        \depedge[edge style={blue!60!}, edge below]{2}{11}{.}
        \depedge[edge style={red!60!}, edge below]{4}{10}{.}
        \depedge[edge style={blue!60!}, edge below]{8}{10}{.}
        \end{dependency}
        
        \hspace{3.8cm}
        \bf BERT
    \end{minipage}
    \hspace{1.5cm}
    \begin{minipage}[b]{0.4\textwidth}
        \begin{dependency}[hide label, edge unit distance=.5ex]
          \begin{deptext}[column sep=0.05cm]
          We\& land\& and\& spill\& out\& and\& go\& our\& separate\& ways\& . \\
        \end{deptext}
        \depedge{1}{2}{.}
        \depedge{2}{4}{.}
        \depedge{2}{7}{.}
        \depedge{2}{11}{.}
        \depedge{3}{4}{.}
        \depedge{4}{5}{.}
        \depedge{6}{7}{.}
        \depedge{7}{10}{.}
        \depedge{8}{10}{.}
        \depedge{9}{10}{.}
        \depedge[edge style={blue!60!}, edge below]{2}{4}{.}
        \depedge[edge style={blue!60!}, edge below]{7}{10}{.}
        \depedge[edge style={blue!60!}, edge below]{2}{11}{.}
        \depedge[edge style={blue!60!}, edge below]{1}{2}{.}
        \depedge[edge style={blue!60!}, edge below]{6}{7}{.}
        \depedge[edge style={blue!60!}, edge below]{4}{5}{.}
        \depedge[edge style={blue!60!}, edge below]{2}{7}{.}
        \depedge[edge style={blue!60!}, edge below]{9}{10}{.}
        \depedge[edge style={blue!60!}, edge below]{3}{4}{.}
        \depedge[edge style={blue!60!}, edge below]{8}{10}{.}
        \end{dependency}
        
        \hspace{3.8cm}
        \bf UDify
    \end{minipage}

    \caption{\label{fig:deps-probe}
        Examples of minimum spanning trees produced by the syntactic probe are shown below each sentence, evaluated on BERT (left) and on UDify (right). Gold dependency trees are shown above each sentence in black. Matched and unmatched spanning tree edges are shown in blue and red respectively.
    }
\end{figure*}

Echoing results seen in \newcite{smith201882}, UDify also shows strong improvement leveraging multilingual data from other UD treebanks.
In low-resource cases, fine-tuning BERT on all treebanks can be far superior to fine-tuning monolingually.
A second round of fine-tuning on an individual treebank using UDify's BERT weights can improve this further, especially for treebanks that underperform the baseline.
However, for languages that are already display strong results, we typically notice worse evaluation performance across all the evaluation metrics.
This indicates that multilingual fine-tuning really is superior to single language fine-tuning with respect to these high-performing languages, showing improvements of up to 20\% reduction in error.

Interestingly, Slavic languages tend to perform the best with multilingual training.
While languages like Czech and Russian possess the largest UD treebanks and do not differ as much in performance from monolingual fine-tuning, evidenced by the improvements over single-language fine-tuning, we can see a large degree of morphological and syntactic structure has transferred to low-resource Slavic languages like Upper Sorbian, whose treebank contains only 646 sentences.
But this is not only true of Slavic languages, as the Turkic language Kazakh (with less than 1,000 training sentences) has also improved significantly.

The zero-shot results indicate that fine-tuning on BERT can result in reasonably high scores on languages that do not have a training set.
It can be seen that a combination of BERT pretraining and multilingual learning can improve predictions for Breton and Tagalog, which implies that the network has learned representations of syntax that cross lingual boundaries.
Furthermore, despite the fact that neither BERT nor UDify have directly observed Faroese, Naija, or Sanskrit, we see unusually high performance in these languages.
This can be partially attributed to each language closely resembling another: Faroese is very close to Icelandic, Naija (Nigerian Pidgin) is a variant of English, and Sanskrit is an ancient Indian language related to Greek, Latin, and Hindi.

Table~\ref{table:results-ablation} shows that layer attention on BERT for each task is beneficial for test performance, much more than using a global weighted average.
In fact, Figure~\ref{fig:bert-layer-weights} shows that each task prefers the layers of BERT differently, uniquely extracting the optimal information for a task.
All tasks favor the information content in the last 3 layers, with a tendency to disprefer layers closer to the input.
However, an interesting observation is that for Lemmas and UFeats, the classifier prefers to also incorporate the information of the first 3 layers.
This meshes well with the linguistic intuition that morphological features are more closely related to the surface form of a word and rely less on context than other syntactic tasks.
Curiously enough, the middle layers are highly dispreferred, meaning that the most useful processing for multilingual syntax (tagging, dependency parsing) occurs in the last 3-4 layers.
The results released by \newcite{tenney2019bert} also agree with the intuition behind the weight distribution above, showing how the different layers of BERT generate hierarchical information like a traditional NLP pipeline, starting with low-level syntax (e.g., POS tagging) and building up to high-level syntactic and semantic dependency parsing.

\subsection{Effect of Syntactic Fine-Tuning on BERT}

Even without any supervised training, BERT encodes its syntax in the embedding's distance close to human-annotated dependencies. But more notably, the results in Table~\ref{table:probe-comparison} show that fine-tuning BERT on Universal Dependencies significantly boosts UUAS scores when compared to the gold dependency trees, an error reduction of 41\%.
This indicates that the self-attention weights have learned a linearly-transformable representation of its vectors more closely resembling annotated dependency trees defined by linguists.
Even with just unsupervised pretraining, a global structural property of the vector space of the BERT weights already produces a decent representation of the dependency tree in the squared L2 distance.
Following this, it should be no surprise that training with a non-linear graph-based dependency decoder would produce even higher quality dependency trees.

% This also demonstrates why BERT can produce high-quality syntax so early on in fine-tuning with just a few epochs, as shown in \newcite{devlin2018bert}. 

\subsection{Attention Visualization}

\begin{figure*}[htbp]
    \centering
    \includegraphics[height=8.5cm]{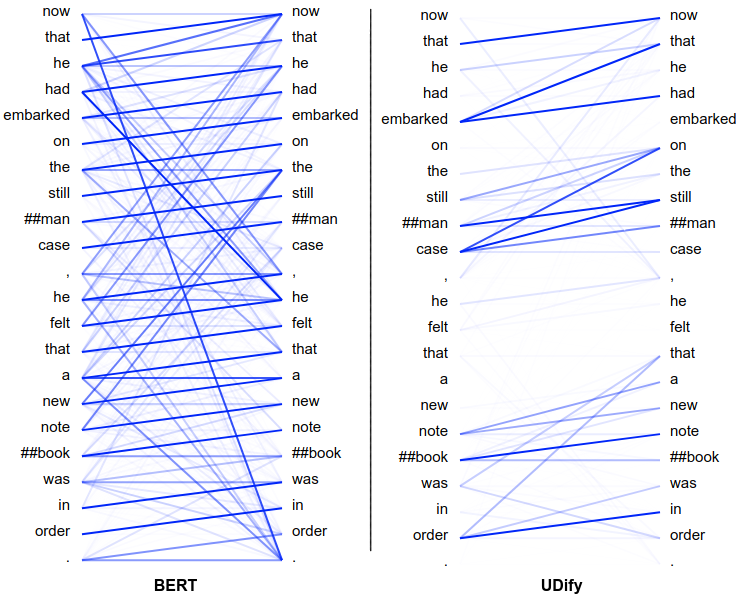}
    \caption{\label{fig:attention-comparison}
        Visualization of BERT attention head 4 at layer 11, comparing the attended words on an English sentence between BERT base and UDify BERT after fine-tuning. The right column indicates the attended words (keys) with respect to the words in the left column (queries). Darker lines indicate stronger attention weights.
    }
\end{figure*}

We performed a high-level visual analysis of the BERT attention weights to see if they have changed on any discernible level.
Our observations reveal something notable: the attention weights tend to be more sparse, and are more often sensitive to constituent boundaries like clauses and prepositions.
Figure~\ref{fig:attention-comparison} illustrates this point, showing the attention weights of a particular attention head on an example sentence.
We find similar behavior in 13 additional attention heads for the provided example sentence.

We see that some of the attention structure remains after fine-tuning.
Previously, the attention head was mostly sensitive to previous words and punctuation.
But after fine-tuning, it demonstrates more fine-grained attention towards immediate wordpieces, prepositions, articles, and adjectives.
We found similar evidence in other attention heads, which implies that fine-tuning on UD produces attention that more closely resembles localized dependencies within constituents.
We also find that BERT base heavily preferred to attend to punctuation, while UDify BERT does to a much lesser degree.

\subsection{Factors that Enable BERT to Excel at Dependency Parsing and Multilinguality}

\citet{goldberg2019assessing} assesses the syntactic capabilities of BERT and concludes that BERT is remarkably capable of processing syntactic tasks despite not being trained on any supervised data.
Conducting similar experiments, \citet{vig2019visualizing} and \citet{sileo2019understanding} visualize the attention heads within each BERT layer, showing a number of distinct attention patterns, including attending to previous/next words, related words, punctuation, verbs/nouns, and coreference dependencies.

This neat delegation of certain low-level information processing tasks to the attention heads hints at why BERT might excel at processing syntax.
We see that from the analysis on BERT fine-tuned with syntax using the syntactic probe and attention visualization, BERT produces a representation that keeps constituents close in its vector space, and improves this representation to more closely resemble human annotated dependency trees when fine-tuned on UD as seen in Figure~\ref{fig:deps-probe}.
Furthermore, \newcite{ahmad2018near} provide results consistent with their claim that self-attention networks can be more robust than recurrent networks to the change of word order, observing that self-attention networks capture less word order information in their architecture, which is what allows them to generally perform better at cross-lingual parsing.
\newcite{wu2019beto} also analyze multilingual BERT and report that the model retains both language-independent as well as language-specific information related to each input sentence, and that the shared embedding space with the input wordpieces correlates strongly with cross-lingual generalization.

From the evidence above, we can see that the combination of strong regularization paired with the ability to capture long-range dependencies with self-attention and contextual pretraining on an enormous corpus of raw text are large contributors that enable robust multilingual modeling with respect to dependency parsing.
Pretraining self-attention networks introduces a strong syntactic bias that is capable of generalizing across languages.
The dependencies seen in the output dependency trees are highly correlated with the implicit dependencies learned by the self-attention, showing that self-attention is remarkably capable of modeling syntax by picking up on common syntactic patterns in text.
The introduction of multilingual data also shows that these attention heads provide a surprising amount of capacity that do not degrade the performance considerably when compared to monolingual training.
E.g., \newcite{devlin2018bert} report that the fine-tuning on the multilingual BERT model results in a small degradation in English fine-tune performance with 104 pretrained languages compared to an equivalent model pretrained only on English.
This also hints that the BERT model can be compressed significantly without compromising heavily on evaluation performance.

\section{Related Work} \label{sec:related-work}

This work's main contribution in combining treebanks for multilingual UD parsing is most similar to the Uppsala system for the CoNLL 2018 Shared Task \cite{smith201882}.
Uppsala combines treebanks of one language or closely related languages together over 82 treebanks and parses all UD annotations in a multi-task pipeline architecture for a total of 34 models.
This approach reduces the number of models required to parse each language while also showing results that are no worse than training on each treebank individually, and in especially low-resource cases, significantly improved.
Combining UD treebanks in a language-agnostic way was first introduced in \newcite{vilares2016one}, which train bilingual parsers on pairs of UD treebanks, showing similar improvements.

Other efforts in training multilingual models include \newcite{johnson2017google}, which demonstrate a machine translation model capable of supporting translation between 12 languages.
Recurrent models have also shown to be capable of scaling to a larger number of languages as seen in \newcite{artetxe2018massively}, which define a scalable approach to train massively multilingual embeddings using recurrent networks on an auxiliary task, e.g., natural language inference.
\newcite{schuster2019cross} produce context-independent multilingual embeddings using a novel embedding alignment strategy to allow models to improve the use of cross-lingual information, showing improved results in dependency parsing.

% Sharing training sets of similar languages for multilingual parameter sharing can improve not only syntactic tasks like dependency parsing \cite{naseem2012selective, duong2015low, ammar2016many, de2018parameter}, but also more complex tasks like neural machine translation \cite{dong2015multi, firat2016multi, lu2018neural}.

\section{Conclusion}

We have proposed and evaluated UDify, a multilingual multi-task self-attention network fine-tuned on BERT pretrained embeddings, capable of producing annotations for any UD treebank, and exceeding the state-of-the-art in UD dependency parsing in a large subset of languages while being comparable in tagging and lemmatization accuracy.
Strong regularization and task-specific layer attention are highly beneficial for fine-tuning, and coupled with training multilingually, also reduce the number of required models to train down to one.
Multilingual learning is most beneficial for low-resource languages, even ones that do not possess a training set, and can be further improved by fine-tuning monolingually using BERT weights saved from UDify's multilingual training.
All these results indicate that self-attention networks are remarkably capable of capturing syntactic patterns, and coupled with unsupervised pretraining are able to scale to a large number of languages without degrading performance.

\section*{Acknowledgments}

The work described herein has been supported by OP VVV VI LINDAT/CLARIN project
of the Ministry of Education, Youth and Sports of the Czech Republic (project
CZ.02.1.01/0.0/0.0/16\_013/0001781) and it has been supported and has been
using language resources developed by the LINDAT/CLARIN project of the Ministry
of Education, Youth and Sports of the Czech Republic (project LM2015071).

Dan Kondratyuk has been supported by the Erasmus Mundus program in Language \& Communication Technologies (LCT), and by the German Federal Ministry of Education and Research (BMBF) through the project DEEPLEE (01IW17001).

\bibliography{acl2019}
\bibliographystyle{acl_natbib}

\appendix 

\section{Appendix} \label{sec:appendix}

% In this section, we illustrate the main data flow of the UDify architecture in Section~\ref{sec:udify-architecture}, explain some additional hyperparameter and architecture choices and other interesting observations in Section~\ref{sec:misc-details}, highlight related work in Section~\ref{sec:related-work}, and show full tables of results seen in the paper in Section~\ref{sec:full-results}.
In this section, we detail and explain hyperparameter choices and miscellaneous details related to model training and display the full tables of evaluation results of UDify across all UD languages.

% illustrate the main data flow of the UDify architecture in Section~\ref{sec:udify-architecture}, explain some additional hyperparameter and architecture choices and other interesting observations in Section~\ref{sec:misc-details}, and show full tables of results seen in the paper in Section~\ref{sec:full-results}.

\subsection{Hyperparameters} \label{sec:udify-hyperparameters} %%% GRAD STUDENT DESCENT %%% BERT IS NICE AND STRONG , BUT IS IT BACKPROPAGANDA ? %%%

% Figure~\ref{fig:udify-architecture} illustrates an example sentence being processed by UDify.
% An input word-tokenized sentence is tokenized by BERT.
% This sentence is then processed through BERT, producing 12 hidden vectors (one for each layer) of dimension 768 for each wordpiece.
% The vectors whose BERT layers correspond to the first wordpiece of each word are combined together using layer attention, producing 4 vectors per token, one for each of the 4 tasks. 
% These are independently processed through dense projection and softmax operations, which are finally decoded to their respective output values.

% The UPOS, UFeats, and Lemma tasks use simple projection layers for sequence tagging, while the dependency parser produces two sets of intermediate projections corresponding to tag and arc dependency predictions, which are combined together using biaffine matrix operations to calculate scores across all possible head-relation pairs.
% See \newcite{dozat2016deep} for more details.

\begin{table}[htbp]
    % \fontsize{8}{10}\selectfont
    \small
    \begin{center}
    \begin{tabular}{@{}lr@{}}
    \toprule
        \sc Hyperparameter & \sc Value \\
    \midrule
        Dependency tag dimension       & 256 \\
        Dependency arc dimension       & 768 \\
        Optimizer          & Adam \\
        $\beta_1,\beta_2$  & 0.9, 0.99 \\
        Weight decay       & 0.01 \\
        Label Smoothing    & 0.03 \\
        Dropout            & 0.5 \\
        BERT dropout       & 0.2 \\
        Mask probability   & 0.2 \\
        Layer dropout      & 0.1 \\
        Batch size         & 32 \\
        Epochs             & 80 \\
        Base learning rate            & $1e^{-3}$ \\
        BERT learning rate            & $5e^{-5}$ \\
        Learning rate warmup steps    & 8000 \\
        Gradient clipping  & 5.0 \\
    \bottomrule
    \end{tabular}
    \end{center}
    \caption{\label{table:hyperparameters}
        A summary of model hyperparameters.
    }
\end{table}

Upon concatenating all training sets, we shuffle all the sentences, bundle them into batches of 32 sentences each, and train UDify for a total of 80 epochs before stopping.
We hold the learning rate constant until we unfreeze BERT in the second epoch, where we and linearly warm up the learning rate for the next 8,000 batches and then apply inverse square root learning rate decay for the remaining epochs.
For the dependency parser, we use feedforward tag and arc dimensions of 300 and 800 respectively.
We apply a small weight decay penalty of 0.01 to ensure that the weights remain small after each update.
For optimization we use the Adam optimizer and we compute softmax cross entropy loss to train the network.
We use a default $\beta_1$ value of 0.9 and lower the $\beta_2$ value from the typical 0.999 to 0.99. The reasoning is to increase the decay rate of the second moment in the Adam optimizer to reduce the chance of the optimizer being too optimistic with respect to the gradient history.
We clip the gradient updates to a maximum L2 magnitude of 5.0. A summary of hyperparameters can be found in Table~\ref{table:hyperparameters}.

To speed up training, we employ bucketed batching, sorting all sentences by their length and grouping similar length sentences into each batch.
However, to ensure that most sentences do not get grouped within the same batch, we fuzz the lengths of each sentence by a maximum of 10\% of its true length when grouping sentences together.

Despite using all the regularization strategies shown previously, we still observe overfitting and must apply more aggressive techniques.
To further regularize the network, we also increase the attention and hidden dropout rates of BERT from 0.1 to 0.2, and we also apply a dropout rate of 0.5 to all BERT layers before computing layer attention for each of the four tasks and applying a layer dropout with probability 0.1.
We increase the masking probability of each wordpiece from 0.15 to 0.2.

With all these regularization strategies and hyperparameter choices combined, we are able to fine-tune BERT for far more epochs before the network starts to overfit, i.e., 80 as opposed to around 10. Even so, we believe even more regularization can improve test performance.

The final multilingual UDify model was trained over approximately 25 days on an NVIDIA GTX 1080 Ti taking an average of 8 hours per epoch.
We use half-precision (fp16) training to be able to keep the BERT model in memory.
One notable aspect of training is that while we observed the model start to level out in validation performance at around epoch 30, the model continually made small, incremental improvements over each subsequent epoch, resulting in far higher scores than if the model training was terminated early. This can be partially attributed to the decaying inverse square root learning rate.

Due to the high training times, we are only able to report on a small number of training experiments for the most relevant and useful results. Prior to developing the final model, we conducted fine-tuning experiments on pairs of languages to find a set of hyperparameters that worked best for multilingual learning.
After this, we gradually scaled up training to 3 languages, 5 languages, 15 languages, and then finally the model presented above. We had high doubts, and wanted to see where the limit was in multilingual training.
We were pleasantly surprised to find that this simple training scheme was able to scale up so well to all UD treebanks.

\subsection{Training Size Effect on Performance}

\begin{figure}[tbp]
    \centering
    \includegraphics[width=\linewidth]{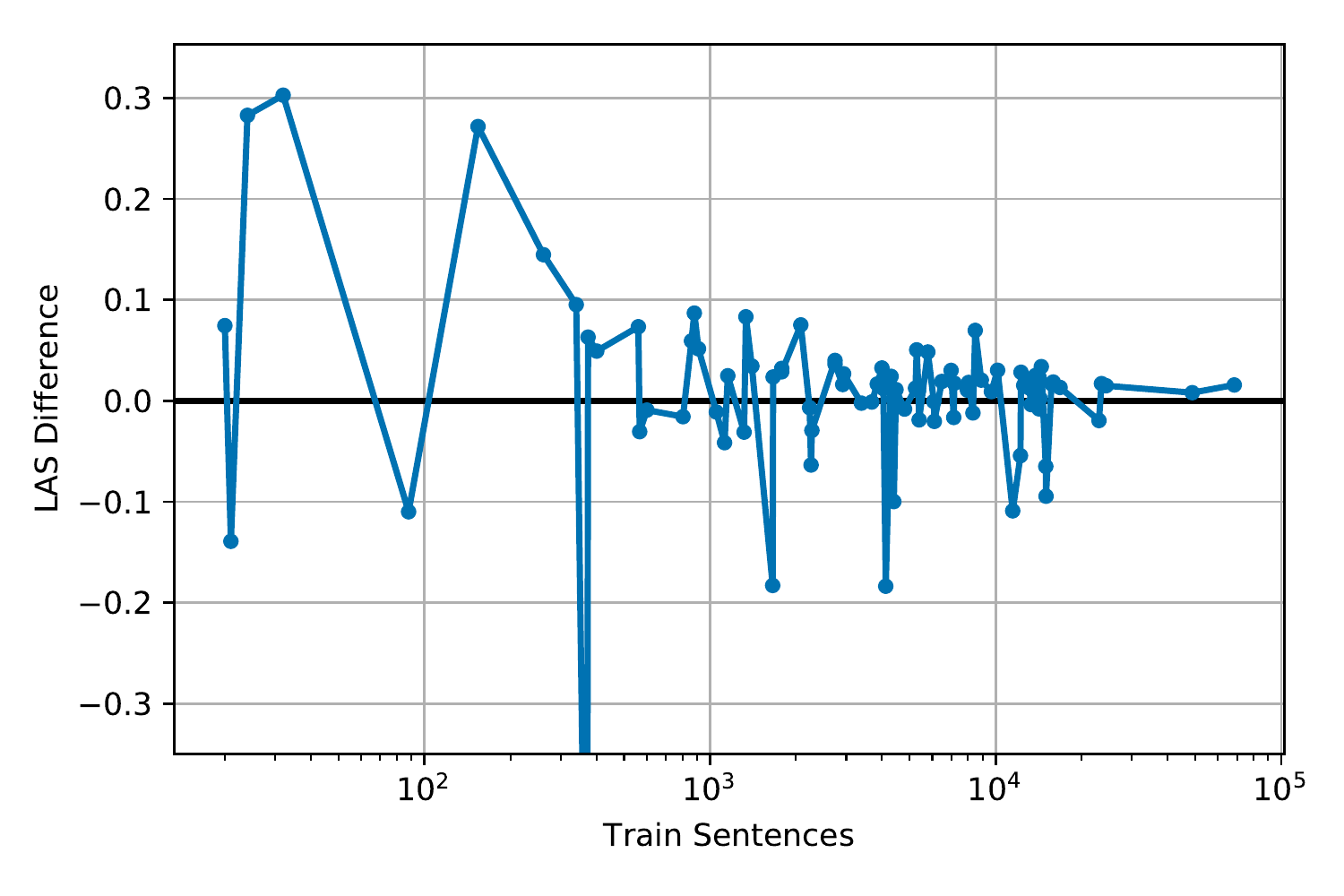}
    \caption{\label{fig:las-comparison}
        A plot of the difference in LAS between UDify and UDPipe Future with respect to the number of training sentences in each treebank.
    }
\end{figure}

\begin{figure}[tbp]
    \centering
    \includegraphics[width=\linewidth]{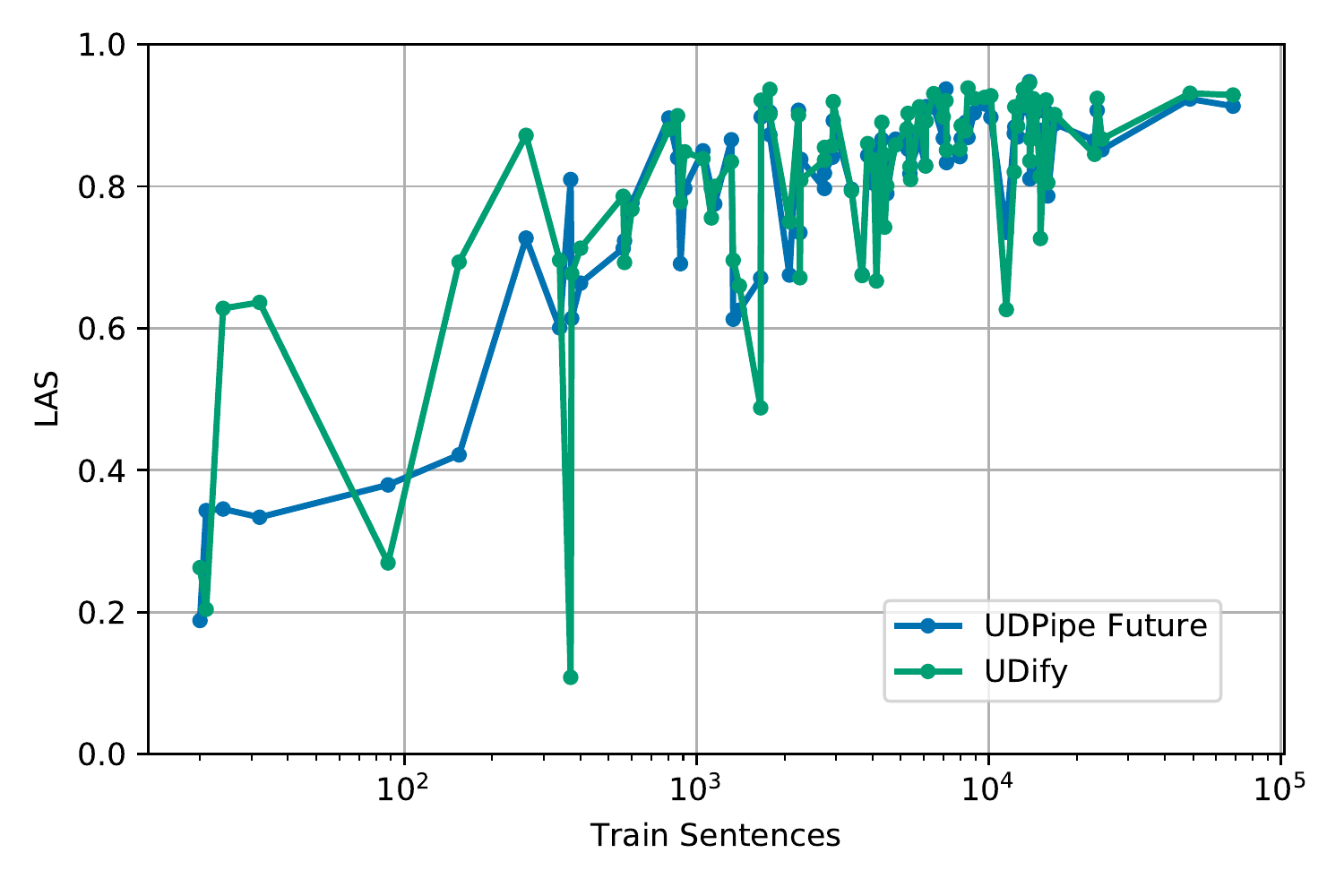}
    \caption{\label{fig:udify-las}
        A plot of LAS between with respect to the number of training sentences in each treebank.
    }
\end{figure}

To gain a better understanding of where the largest score improvements in UDify occur, we plot the LAS improvement UDify provides over UDPipe Future for each treebank, ordered by the size (number of sentences) of the training set, see Figure~\ref{fig:las-comparison}. The results show that the largest improvements tend to occur on small treebanks with less than 3,000 training examples. For absolute LAS values, see Figure~\ref{fig:udify-las}, which indicates that more training resources tend to improve evaluation performance overall.

\subsection{Miscellaneous Details} \label{sec:misc-details}

Our results show that modeling language-specific properties is not strictly necessary to achieve high-performing cross-lingual representations for dependency parsing, though we caution that the model can also likely be improved by these techniques.

Fine-tuning BERT on UD introduces a syntactic bias in the network, and we are interested in observing any differences in transfer learning by fine-tuning this new ``UD-BERT'' on other tasks. We leave a comprehensive evaluation of injecting syntactic bias into language models with respect to knowledge transfer for future work.

We note that saving the weights of BERT and fine-tuning a second round can improve performance as demonstrated in Stickland et al. (2019).
% We note that saving the weights of BERT and fine-tuning a second round can improve performance as demonstrated in \newcite{stickland2019bert}.
The improvements of UDify+Lang over just UDify can be partially attributed to this, but we can see that even these improvements can be inferior to fine-tuning on all UD treebanks.

BERT limits its positional encoding to 512 wordpieces, causing some sentences in UD to be too long to fit into the model.
We use a sliding window approach to break up long sentences into windows of 512 wordpieces, overlapping each window by 256 wordpieces.
After feeding the windows into BERT, we select the first 256 wordpieces of each window and any remaining wordpieces in the last window to represent the contextual embeddings of each word in the original sentence.

% \subsection{Penn Treebank}

% To test the strength of our fine-tuning approach, we apply the same UDify training scheme on the Penn Treebank. We also show the effect of performing an intermediate step . From the results in Table~\ref{}, we achieve state-of-the-art results in dependency parsing on the Penn Treebank. We converted the Penn Treebank files to Stanford Dependencies (version 3.3.0) to match the same formatting as CVT.

% BERT Large Finetune x1
% POS 9
% UAS 96.49
% LAS 95.01

% BERT Large Finetune x2
% POS 97.95
% UAS 96.67
% LAS 95.12

\subsection{Full Results of UD Scores} \label{sec:full-results}

We show in Tables~\ref{table:full-results-1},~\ref{table:full-results-2},~\ref{table:full-results-3}, and~\ref{table:full-results-4} UDify scores evaluated on all 124 treebanks with the official CoNLL 2018 Shared Task evaluation script.
For comparison, we also include the full test evaluation of UDPipe Future on the subset of 89 treebanks with a training set.
We also add a column indicating the size of each treebank, i.e., the number of sentences in the training set.
% We also make a machine-readable version of each table with higher numerical precision available in the repository code.

\begin{table*}
    \fontsize{8}{10}\selectfont
    \begin{center}
    \begin{tabular}{@{}llrrrrrrrrr@{}}
    \toprule
    \sc Treebank & \sc Model & \sc UPOS & \sc UFeats & \sc Lemmas & \sc UAS & \sc LAS & \sc CLAS & \sc  MLAS & \sc BLEX & \sc Size \\
    \midrule
    Afrikaans AfriBooms & UDPipe &  \bf 98.25 &  \bf 97.66 &   \bf 97.46 &  \bf 89.38 &  \bf 86.58 &  \bf 81.44 &  \bf 77.66 &  \bf 77.82 &   1.3k \\
               & UDify &      97.48 &      96.63 &       95.23 &      86.97 &      83.48 &      77.42 &      70.57 &      70.93 &   1.3k \\
    \addlinespace
    Akkadian PISANDUB & UDify &  \bf 19.92 &  \bf 99.51 &    \bf 2.32 &  \bf 27.65 &   \bf 4.54 &   \bf 3.27 &   \bf 1.04 &   \bf 0.30 &      0 \\
    \addlinespace
    Amharic ATT & UDify &  \bf 15.25 &  \bf 43.95 &   \bf 58.04 &  \bf 17.38 &   \bf 3.49 &   \bf 4.88 &   \bf 0.23 &   \bf 2.53 &      0 \\
    \addlinespace
    Ancient Greek PROIEL & UDPipe &  \bf 97.86 &  \bf 92.44 &   \bf 93.51 &  \bf 85.93 &  \bf 82.11 &  \bf 77.70 &  \bf 67.16 &  \bf 71.22 &  15.0k \\
               & UDify &      91.20 &      82.29 &       76.16 &      78.91 &      72.66 &      66.07 &      50.79 &      47.27 &  15.0k \\
    \addlinespace
    Ancient Greek Perseus & UDPipe &  \bf 93.27 &  \bf 91.39 &   \bf 85.02 &  \bf 78.85 &  \bf 73.54 &  \bf 67.60 &  \bf 53.87 &  \bf 53.19 &  11.5k \\
               & UDify &      85.67 &      81.67 &       70.51 &      70.51 &      62.64 &      55.60 &      39.15 &      35.05 &  11.5k \\
    \addlinespace
    Arabic PADT & UDPipe &  \bf 96.83 &  \bf 94.11 &   \bf 95.28 &      87.54 &  \bf 82.94 &  \bf 79.77 &  \bf 73.92 &  \bf 75.87 &   6.1k \\
               & UDify &      96.58 &      91.77 &       73.55 &  \bf 87.72 &      82.88 &      79.47 &      70.52 &      50.26 &   6.1k \\
    \addlinespace
    Arabic PUD & UDify &  \bf 79.98 &  \bf 40.32 &    \bf 0.00 &  \bf 76.17 &  \bf 67.07 &  \bf 65.10 &  \bf 10.67 &   \bf 0.00 &      0 \\
    \addlinespace
    Armenian ArmTDP & UDPipe &      93.49 &  \bf 82.85 &   \bf 92.86 &      78.62 &      71.27 &      65.77 &  \bf 48.11 &  \bf 60.11 &    561 \\
               & UDify &  \bf 94.42 &      76.90 &       85.63 &  \bf 85.63 &  \bf 78.61 &  \bf 73.72 &      46.80 &      59.14 &    561 \\
    \addlinespace
    Bambara CRB & UDify &  \bf 30.86 &  \bf 57.96 &   \bf 20.42 &  \bf 30.28 &   \bf 8.60 &   \bf 6.56 &   \bf 1.04 &   \bf 0.76 &      0 \\
    \addlinespace
    Basque BDT & UDPipe &  \bf 96.11 &  \bf 92.48 &   \bf 96.29 &  \bf 86.11 &  \bf 82.86 &  \bf 81.79 &  \bf 72.33 &  \bf 78.54 &   5.4k \\
               & UDify &      95.45 &      86.80 &       90.53 &      84.94 &      80.97 &      79.52 &      63.60 &      71.56 &   5.4k \\
    \addlinespace
    Belarusian HSE & UDPipe &      93.63 &      73.30 &   \bf 87.34 &      78.58 &      72.72 &      69.14 &      46.20 &      58.28 &    261 \\
               & UDify &  \bf 97.54 &  \bf 89.36 &       85.46 &  \bf 91.82 &  \bf 87.19 &  \bf 85.05 &  \bf 71.54 &  \bf 68.66 &    261 \\
    \addlinespace
    Breton KEB & UDify &  \bf 62.78 &  \bf 47.12 &   \bf 51.31 &  \bf 63.52 &  \bf 39.84 &  \bf 35.14 &   \bf 4.64 &  \bf 16.34 &      0 \\
    \addlinespace
    Bulgarian BTB & UDPipe &  \bf 98.98 &  \bf 97.82 &   \bf 97.94 &      93.38 &      90.35 &      87.01 &  \bf 83.63 &  \bf 84.42 &   8.9k \\
               & UDify &      98.89 &      96.18 &       93.49 &  \bf 95.54 &  \bf 92.40 &  \bf 89.59 &      83.43 &      80.44 &   8.9k \\
    \addlinespace
    Buryat BDT & UDPipe &      40.34 &      32.40 &       58.17 &      32.60 &      18.83 &      12.36 &       1.26 &   \bf 6.49 &     20 \\
               & UDify &  \bf 61.73 &  \bf 47.45 &   \bf 61.03 &  \bf 48.43 &  \bf 26.28 &  \bf 20.61 &   \bf 5.51 &      11.68 &     20 \\
    \addlinespace
    Cantonese HK & UDify &  \bf 67.11 &  \bf 91.01 &   \bf 96.01 &  \bf 46.82 &  \bf 32.01 &  \bf 33.35 &  \bf 14.29 &  \bf 31.26 &      0 \\
    \addlinespace
    Catalan AnCora & UDPipe &      98.88 &  \bf 98.37 &   \bf 99.07 &      93.22 &      91.06 &      87.18 &      84.48 &      86.18 &  13.1k \\
           & UDify &  \bf 98.89 &      98.34 &       98.14 &  \bf 94.25 &  \bf 92.33 &  \bf 89.27 &  \bf 86.21 &  \bf 86.61 &  13.1k \\
    \addlinespace
    Chinese CFL & UDify &  \bf 83.75 &  \bf 82.72 &   \bf 98.75 &  \bf 62.46 &  \bf 42.48 &  \bf 43.46 &  \bf 21.07 &  \bf 42.22 &      0 \\
    \addlinespace
    Chinese GSD & UDPipe &      94.88 &      99.22 &   \bf 99.99 &      84.64 &      80.50 &      76.79 &      71.04 &      76.78 &   4.0k \\
               & UDify &  \bf 95.35 &  \bf 99.35 &       99.97 &  \bf 87.93 &  \bf 83.75 &  \bf 80.33 &  \bf 74.36 &  \bf 80.28 &   4.0k \\
    \addlinespace
    Chinese HK & UDify &  \bf 82.86 &  \bf 86.47 &  \bf 100.00 &  \bf 65.53 &  \bf 49.32 &  \bf 47.84 &  \bf 22.85 &  \bf 47.84 &      0 \\
    \addlinespace
    Chinese PUD & UDify &  \bf 92.68 &  \bf 98.40 &  \bf 100.00 &  \bf 79.08 &  \bf 56.51 &  \bf 55.22 &  \bf 40.92 &  \bf 55.22 &      0 \\
    \addlinespace
    Coptic Scriptorium & UDPipe &  \bf 94.70 &  \bf 96.35 &   \bf 95.49 &  \bf 85.58 &  \bf 80.97 &  \bf 72.24 &  \bf 64.45 &  \bf 68.48 &    371 \\
               & UDify &      27.17 &      52.85 &       55.71 &      27.58 &      10.82 &       6.50 &       0.19 &       1.44 &    371 \\
    \addlinespace
    Croatian SET & UDPipe &  \bf 98.13 &  \bf 92.25 &   \bf 97.27 &      91.10 &      86.78 &      84.11 &  \bf 73.61 &      81.19 &   7.0k \\
               & UDify &      98.02 &      89.67 &       95.34 &  \bf 94.08 &  \bf 89.79 &  \bf 87.70 &      72.72 &  \bf 82.00 &   7.0k \\
    \addlinespace
    Czech CAC & UDPipe &  \bf 99.37 &  \bf 96.34 &   \bf 98.57 &      92.99 &      90.71 &      88.84 &      84.30 &      87.18 &  23.5k \\
               & UDify &      99.14 &      95.42 &       98.32 &  \bf 94.33 &  \bf 92.41 &  \bf 91.03 &  \bf 84.68 &  \bf 89.21 &  23.5k \\
    \addlinespace
    Czech CLTT & UDPipe &      98.88 &      91.59 &       98.25 &      86.90 &      84.03 &      80.55 &      71.63 &      79.20 &    861 \\
               & UDify &  \bf 99.17 &  \bf 93.66 &   \bf 98.86 &  \bf 91.69 &  \bf 89.96 &  \bf 87.59 &  \bf 79.50 &  \bf 86.79 &    861 \\
    \addlinespace
    Czech FicTree & UDPipe &  \bf 98.55 &  \bf 95.87 &   \bf 98.63 &      92.91 &      89.75 &      86.97 &  \bf 81.04 &      85.49 &  10.2k \\
               & UDify &      98.34 &      91.82 &       98.13 &  \bf 95.19 &  \bf 92.77 &  \bf 90.99 &      77.77 &  \bf 88.39 &  10.2k \\
    \addlinespace
    Czech PDT & UDPipe &  \bf 99.18 &  \bf 97.23 &   \bf 99.02 &      93.33 &      91.31 &      89.64 &      86.15 &      88.60 &  68.5k \\
               & UDify &  \bf 99.18 &      96.69 &       98.52 &  \bf 94.73 &  \bf 92.88 &  \bf 91.64 &  \bf 87.13 &  \bf 89.95 &  68.5k \\
    \addlinespace
    Czech PUD & UDify &  \bf 97.93 &  \bf 93.98 &   \bf 96.94 &  \bf 92.59 &  \bf 87.95 &  \bf 84.85 &  \bf 77.39 &  \bf 82.81 &      0 \\
    \addlinespace
    Danish DDT & UDPipe &  \bf 97.78 &  \bf 97.33 &   \bf 97.52 &      86.88 &      84.31 &      81.20 &  \bf 76.29 &  \bf 78.51 &   4.4k \\
               & UDify &      97.50 &      95.41 &       94.60 &  \bf 87.76 &  \bf 84.50 &  \bf 81.60 &      73.76 &      75.15 &   4.4k \\
    \addlinespace
    Dutch Alpino & UDPipe &      96.83 &      96.33 &   \bf 97.09 &      91.37 &      88.38 &      83.51 &      77.28 &      79.82 &  12.3k \\
               & UDify &  \bf 97.67 &  \bf 97.66 &       95.44 &  \bf 94.23 &  \bf 91.21 &  \bf 87.32 &  \bf 82.81 &  \bf 80.76 &  12.3k \\
    \addlinespace
    Dutch LassySmall & UDPipe &      96.50 &      96.42 &   \bf 97.41 &      90.20 &      86.39 &      81.88 &      77.19 &      78.83 &   5.8k \\
               & UDify &  \bf 96.70 &  \bf 96.57 &       95.10 &  \bf 94.34 &  \bf 91.22 &  \bf 88.03 &  \bf 82.06 &  \bf 81.40 &   5.8k \\
    \bottomrule
    \end{tabular}
    \end{center}
    \caption{\label{table:full-results-1} The full test results of UDify on 124 treebanks (part 1 of 4). The {\sc Size} column indicates the number of training sentences.}
\end{table*}

\begin{table*}
    \fontsize{8}{10}\selectfont
    \begin{center}
    \begin{tabular}{@{}llrrrrrrrrr@{}}
    \toprule
    \sc Treebank & \sc Model & \sc UPOS & \sc UFeats & \sc Lemmas & \sc UAS & \sc LAS & \sc CLAS & \sc  MLAS & \sc BLEX & \sc Size \\
    \midrule
    English EWT & UDPipe &  \bf 96.29 &  \bf 97.10 &   \bf 98.25 &      89.63 &      86.97 &      84.02 &      79.00 &      82.36 &  12.5k \\
               & UDify &      96.21 &      96.02 &       97.28 &  \bf 90.96 &  \bf 88.50 &  \bf 86.25 &  \bf 79.80 &  \bf 83.39 &  12.5k \\
    \addlinespace   
    English GUM & UDPipe &  \bf 96.02 &  \bf 96.82 &   \bf 96.85 &      87.27 &      84.12 &      78.55 &  \bf 73.51 &  \bf 74.68 &   2.9k \\
               & UDify &      95.44 &      94.12 &       93.15 &  \bf 89.14 &  \bf 85.73 &  \bf 83.03 &      72.55 &      74.30 &   2.9k \\
    \addlinespace
    English LinES & UDPipe &  \bf 96.91 &  \bf 96.31 &   \bf 96.45 &      84.15 &      79.71 &      77.44 &  \bf 71.38 &      73.22 &   2.7k \\
               & UDify &      95.31 &      91.34 &       94.50 &  \bf 87.33 &  \bf 83.71 &  \bf 82.95 &      68.62 &  \bf 76.23 &   2.7k \\
    \addlinespace
    English PUD & UDify &  \bf 96.18 &  \bf 93.50 &   \bf 94.20 &  \bf 91.52 &  \bf 88.66 &  \bf 87.83 &  \bf 75.61 &  \bf 80.57 &      0 \\
    \addlinespace
    English ParTUT & UDPipe &      96.10 &  \bf 95.51 &   \bf 97.74 &      90.29 &      87.27 &      82.58 &  \bf 76.44 &      80.33 &   1.8k \\
               & UDify &  \bf 96.16 &      92.61 &       96.45 &  \bf 92.84 &  \bf 90.14 &  \bf 86.28 &      74.59 &  \bf 82.01 &   1.8k \\
    \addlinespace
    Erzya JR & UDify &  \bf 46.66 &  \bf 31.82 &   \bf 45.73 &  \bf 31.90 &  \bf 16.38 &  \bf 10.83 &   \bf 0.58 &   \bf 2.83 &      0 \\
    \addlinespace
    Estonian EDT & UDPipe &  \bf 97.64 &  \bf 96.23 &   \bf 95.30 &      88.00 &      85.18 &      83.62 &      78.72 &  \bf 78.51 &  24.4k \\
               & UDify &      97.44 &      95.13 &       86.56 &  \bf 89.53 &  \bf 86.67 &  \bf 85.17 &  \bf 79.20 &      69.31 &  24.4k \\
    \addlinespace
    Faroese OFT & UDify &  \bf 77.46 &  \bf 35.20 &   \bf 51.09 &  \bf 67.24 &  \bf 59.26 &  \bf 51.17 &   \bf 2.39 &  \bf 21.92 &      0 \\
    \addlinespace
    Finnish FTB & UDPipe &  \bf 96.65 &  \bf 96.62 &   \bf 95.49 &  \bf 90.68 &  \bf 87.89 &  \bf 85.11 &  \bf 80.58 &  \bf 81.18 &  15.0k \\
               & UDify &      93.80 &      90.38 &       88.80 &      86.37 &      81.40 &      81.01 &      68.16 &      70.15 &  15.0k \\
    \addlinespace
    Finnish PUD & UDify &  \bf 96.48 &  \bf 93.84 &   \bf 84.64 &  \bf 89.76 &  \bf 86.58 &  \bf 86.64 &  \bf 77.83 &  \bf 69.12 &      0 \\
    \addlinespace
    Finnish TDT & UDPipe &  \bf 97.45 &  \bf 95.43 &   \bf 91.45 &  \bf 89.88 &  \bf 87.46 &  \bf 85.87 &  \bf 80.43 &  \bf 76.64 &  12.2k \\
               & UDify &      94.43 &      90.48 &       82.89 &      86.42 &      82.03 &      82.62 &      70.89 &      63.66 &  12.2k \\
    \addlinespace
    French GSD & UDPipe &      97.63 &  \bf 97.13 &   \bf 98.35 &      90.65 &      88.06 &      84.35 &      79.76 &      82.39 &  14.5k \\
               & UDify &  \bf 97.83 &      96.17 &       97.34 &  \bf 93.60 &  \bf 91.45 &  \bf 88.54 &  \bf 81.61 &  \bf 84.51 &  14.5k \\
    \addlinespace
    French PUD & UDify &  \bf 91.67 &  \bf 59.65 &  \bf 100.00 &  \bf 88.36 &  \bf 82.76 &  \bf 81.74 &  \bf 25.24 &  \bf 81.74 &      0 \\
    \addlinespace
    French ParTUT & UDPipe &  \bf 96.93 &  \bf 94.43 &   \bf 95.70 &  \bf 92.17 &  \bf 89.63 &  \bf 84.62 &  \bf 75.22 &  \bf 78.07 &    804 \\
               & UDify &      96.12 &      88.36 &       93.97 &      90.55 &      88.06 &      83.19 &      63.03 &      74.03 &    804 \\
    \addlinespace
    French Sequoia & UDPipe &  \bf 98.79 &  \bf 98.09 &   \bf 98.57 &      92.37 &  \bf 90.73 &  \bf 87.55 &  \bf 84.51 &  \bf 85.93 &   2.2k \\
               & UDify &      97.89 &      88.97 &       97.15 &  \bf 92.53 &      90.05 &      86.67 &      67.98 &      82.52 &   2.2k \\
    \addlinespace
    French Spoken & UDPipe &      95.91 &     100.00 &   \bf 96.92 &      82.90 &      77.53 &      71.82 &      68.24 &      69.47 &   1.2k \\
               & UDify &  \bf 96.23 &  \bf 98.67 &       96.59 &  \bf 85.24 &  \bf 80.01 &  \bf 75.40 &  \bf 69.74 &  \bf 72.77 &   1.2k \\
    \addlinespace
    Galician CTG & UDPipe &  \bf 97.84 &  \bf 99.83 &   \bf 98.58 &  \bf 86.44 &  \bf 83.82 &  \bf 78.58 &  \bf 72.46 &  \bf 77.21 &   2.3k \\
               & UDify &      96.51 &      97.10 &       97.08 &      84.75 &      80.89 &      74.62 &      65.86 &      72.17 &   2.3k \\
    \addlinespace
    Galician TreeGal & UDPipe &  \bf 95.82 &  \bf 93.96 &   \bf 97.06 &      82.72 &  \bf 77.69 &      71.69 &  \bf 63.73 &  \bf 68.89 &    601 \\
               & UDify &      94.59 &      80.67 &       94.93 &  \bf 84.08 &      76.77 &  \bf 73.06 &      49.76 &      66.99 &    601 \\
    \addlinespace
    German GSD & UDPipe &      94.48 &  \bf 90.68 &   \bf 96.80 &      85.53 &      81.07 &      76.26 &      58.82 &      72.13 &  13.8k \\
               & UDify &  \bf 94.55 &      90.43 &       94.42 &  \bf 87.81 &  \bf 83.59 &  \bf 80.03 &  \bf 61.27 &  \bf 72.48 &  13.8k \\
    \addlinespace
    German PUD & UDify &  \bf 89.49 &  \bf 30.66 &   \bf 94.77 &  \bf 89.86 &  \bf 84.46 &  \bf 80.50 &   \bf 2.10 &  \bf 72.95 &      0 \\
    \addlinespace
    Gothic PROIEL & UDPipe &  \bf 96.61 &  \bf 90.73 &   \bf 94.75 &      85.28 &  \bf 79.60 &  \bf 76.92 &  \bf 66.70 &  \bf 72.93 &   3.4k \\
               & UDify &      95.55 &      85.97 &       80.57 &  \bf 85.61 &      79.37 &      76.26 &      63.09 &      58.65 &   3.4k \\
    \addlinespace
    Greek GDT & UDPipe &  \bf 97.98 &  \bf 94.96 &   \bf 95.82 &      92.10 &      89.79 &      85.71 &  \bf 78.60 &  \bf 79.72 &   1.7k \\
               & UDify &      97.72 &      93.29 &       89.43 &  \bf 94.33 &  \bf 92.15 &  \bf 88.67 &      77.89 &      71.83 &   1.7k \\
    \addlinespace
    Hebrew HTB & UDPipe &  \bf 97.02 &  \bf 95.87 &   \bf 97.12 &      89.70 &      86.86 &      81.45 &  \bf 75.52 &  \bf 78.14 &   5.2k \\
               & UDify &      96.94 &      93.41 &       94.15 &  \bf 91.63 &  \bf 88.11 &  \bf 83.04 &      72.55 &      74.87 &   5.2k \\
    \addlinespace
    Hindi HDTB & UDPipe &  \bf 97.52 &  \bf 94.15 &   \bf 98.67 &      94.85 &  \bf 91.83 &  \bf 88.21 &  \bf 78.49 &  \bf 86.83 &  13.3k \\
               & UDify &      97.12 &      92.59 &       98.23 &  \bf 95.13 &      91.46 &      87.80 &      75.54 &      86.10 &  13.3k \\
    \addlinespace
    Hindi PUD & UDify &  \bf 87.54 &  \bf 22.81 &  \bf 100.00 &  \bf 71.64 &  \bf 58.42 &  \bf 53.03 &   \bf 3.32 &  \bf 53.03 &      0 \\
    \addlinespace
    Hungarian Szeged & UDPipe &      95.76 &  \bf 91.75 &   \bf 95.05 &      84.04 &      79.73 &      78.65 &  \bf 67.63 &  \bf 73.63 &    911 \\
               & UDify &  \bf 96.36 &      86.16 &       90.19 &  \bf 89.68 &  \bf 84.88 &  \bf 83.93 &      64.27 &      72.21 &    911 \\
    \addlinespace
    Indonesian GSD & UDPipe &  \bf 93.69 &  \bf 95.58 &   \bf 99.64 &      85.31 &      78.99 &      76.76 &  \bf 67.74 &  \bf 76.38 &   4.5k \\
               & UDify &      93.36 &      93.32 &       98.37 &  \bf 86.45 &  \bf 80.10 &  \bf 78.05 &      66.93 &      76.31 &   4.5k \\
    \addlinespace
    Indonesian PUD & UDify &  \bf 76.10 &  \bf 44.23 &  \bf 100.00 &  \bf 77.47 &  \bf 56.90 &  \bf 54.88 &   \bf 7.41 &  \bf 54.88 &      0 \\
    \addlinespace
    Irish IDT & UDPipe &  \bf 92.72 &  \bf 82.43 &   \bf 90.48 &  \bf 80.39 &  \bf 72.34 &  \bf 63.48 &  \bf 46.49 &  \bf 55.32 &    567 \\
               & UDify &      90.49 &      71.84 &       81.27 &      80.05 &      69.28 &      60.02 &      34.39 &      43.07 &    567 \\
    \addlinespace
    Italian ISDT & UDPipe &      98.39 &  \bf 98.11 &   \bf 98.66 &      93.49 &      91.54 &      87.34 &      84.28 &      85.49 &  13.1k \\
               & UDify &  \bf 98.51 &      98.01 &       97.72 &  \bf 95.54 &  \bf 93.69 &  \bf 90.40 &  \bf 86.54 &  \bf 86.70 &  13.1k \\
    \addlinespace
    Italian PUD & UDify &  \bf 94.73 &  \bf 58.16 &   \bf 96.08 &  \bf 94.18 &  \bf 91.76 &  \bf 90.05 &  \bf 25.55 &  \bf 83.74 &      0 \\
    \addlinespace
    Italian ParTUT & UDPipe &  \bf 98.38 &      97.77 &   \bf 98.16 &      92.64 &      90.47 &      85.05 &      81.87 &      82.99 &   1.8k \\
               & UDify &      98.21 &  \bf 98.38 &       97.55 &  \bf 95.96 &  \bf 93.68 &  \bf 89.83 &  \bf 86.83 &  \bf 86.44 &   1.8k \\
    \bottomrule
    \end{tabular}
    \end{center}
    \caption{\label{table:full-results-2} The full test results of UDify on 124 treebanks (part 2 of 4).}
\end{table*}

\begin{table*}
    \fontsize{8}{10}\selectfont
    \begin{center}
    \begin{tabular}{@{}llrrrrrrrrr@{}}
    \toprule
    \sc Treebank & \sc Model & \sc UPOS & \sc UFeats & \sc Lemmas & \sc UAS & \sc LAS & \sc CLAS & \sc  MLAS & \sc BLEX & \sc Size \\
    \midrule
    Japanese GSD & UDPipe &  \bf 98.13 &  \bf 99.98 &   \bf 99.52 &  \bf 95.06 &  \bf 93.73 &  \bf 88.35 &  \bf 86.37 &  \bf 88.04 &   7.1k \\
               & UDify &      97.08 &      99.97 &       98.80 &      94.37 &      92.08 &      86.19 &      82.99 &      85.12 &   7.1k \\
    \addlinespace
    Japanese Modern & UDify &  \bf 74.94 &  \bf 96.14 &   \bf 79.70 &  \bf 74.99 &  \bf 55.62 &  \bf 42.67 &  \bf 30.89 &  \bf 35.47 &      0 \\
    \addlinespace
    Japanese PUD & UDify &  \bf 97.89 &  \bf 99.98 &   \bf 99.31 &  \bf 94.89 &  \bf 93.62 &  \bf 87.92 &  \bf 84.86 &  \bf 87.15 &      0 \\
    \addlinespace
    Kazakh KTB & UDPipe &      55.84 &      40.40 &       63.96 &      53.30 &      33.38 &      27.06 &   \bf 4.82 &      15.10 &     32 \\
               & UDify &  \bf 85.59 &  \bf 65.49 &   \bf 77.18 &  \bf 74.77 &  \bf 63.66 &  \bf 61.84 &      34.23 &  \bf 45.51 &     32 \\
    \addlinespace
    Komi Zyrian IKDP & UDify &  \bf 59.92 &  \bf 39.32 &   \bf 57.56 &  \bf 36.01 &  \bf 22.12 &  \bf 17.45 &   \bf 1.54 &   \bf 6.80 &      0 \\
    \addlinespace
    Komi Zyrian Lattice & UDify &  \bf 38.57 &  \bf 29.45 &   \bf 55.33 &  \bf 28.85 &  \bf 12.99 &  \bf 10.79 &   \bf 0.72 &   \bf 3.28 &      0 \\
    \addlinespace
    Korean GSD & UDPipe &  \bf 96.29 &  \bf 99.77 &   \bf 93.40 &  \bf 87.70 &  \bf 84.24 &  \bf 82.05 &  \bf 79.74 &  \bf 76.35 &   4.4k \\
               & UDify &      90.56 &      99.63 &       82.84 &      82.74 &      74.26 &      71.72 &      65.94 &      57.58 &   4.4k \\
    \addlinespace
    Korean Kaist & UDPipe &  \bf 95.59 &     100.00 &   \bf 94.30 &  \bf 88.42 &  \bf 86.48 &  \bf 84.12 &  \bf 80.72 &  \bf 79.22 &  23.0k \\
               & UDify &      94.67 &  \bf 99.98 &       85.89 &      87.57 &      84.52 &      82.05 &      78.27 &      68.99 &  23.0k \\
    \addlinespace
    Korean PUD & UDify &  \bf 64.43 &  \bf 60.47 &   \bf 70.47 &  \bf 63.57 &  \bf 46.89 &  \bf 45.29 &  \bf 16.26 &  \bf 30.94 &      0 \\
    \addlinespace
    Kurmanji MG & UDPipe &      53.36 &  \bf 41.54 &   \bf 69.58 &  \bf 45.23 &  \bf 34.32 &  \bf 29.41 &   \bf 2.74 &      19.39 &     21 \\
               & UDify &  \bf 60.23 &      37.78 &       58.08 &      35.86 &      20.40 &      14.75 &       1.42 &   \bf 7.28 &     21 \\
    \addlinespace
    Latin ITTB & UDPipe &      98.34 &  \bf 96.97 &   \bf 98.99 &      91.06 &      88.80 &      86.40 &  \bf 82.35 &      85.71 &  16.8k \\
               & UDify &  \bf 98.48 &      95.81 &       98.08 &  \bf 92.43 &  \bf 90.12 &  \bf 87.93 &      82.24 &  \bf 85.97 &  16.8k \\
    \addlinespace
    Latin PROIEL & UDPipe &  \bf 97.01 &  \bf 91.53 &   \bf 96.32 &      83.34 &      78.66 &      76.20 &  \bf 67.40 &  \bf 73.65 &  15.9k \\
               & UDify &      96.79 &      89.49 &       91.79 &  \bf 84.85 &  \bf 80.52 &  \bf 77.96 &      67.18 &      71.00 &  15.9k \\
    \addlinespace
    Latin Perseus & UDPipe &      88.40 &      79.10 &   \bf 81.45 &      71.20 &      61.28 &      56.32 &      41.58 &      45.09 &   1.3k \\
               & UDify &  \bf 90.96 &  \bf 82.09 &       81.08 &  \bf 78.33 &  \bf 69.60 &  \bf 65.95 &  \bf 50.26 &  \bf 51.33 &   1.3k \\
    \addlinespace
    Latvian LVTB & UDPipe &  \bf 96.11 &  \bf 93.01 &   \bf 95.46 &      87.20 &      83.35 &      80.90 &  \bf 71.92 &  \bf 76.64 &   7.2k \\
               & UDify &      96.02 &      89.78 &       91.00 &  \bf 89.33 &  \bf 85.09 &  \bf 82.34 &      69.51 &      72.58 &   7.2k \\
    \addlinespace
    Lithuanian HSE & UDPipe &      81.70 &      60.47 &   \bf 76.89 &      51.98 &      42.17 &      38.93 &      18.17 &      28.70 &    154 \\
               & UDify &  \bf 90.47 &  \bf 70.00 &       67.17 &  \bf 79.06 &  \bf 69.34 &  \bf 66.00 &  \bf 36.21 &  \bf 36.35 &    154 \\
    \addlinespace
    Maltese MUDT & UDPipe &  \bf 95.99 &     100.00 &  \bf 100.00 &  \bf 84.65 &  \bf 79.71 &  \bf 71.49 &  \bf 66.75 &  \bf 71.49 &   1.1k \\
               & UDify &      91.98 &  \bf 99.89 &  \bf 100.00 &      83.07 &      75.56 &      65.08 &      58.14 &      65.08 &   1.1k \\
    \addlinespace
    Marathi UFAL & UDPipe &      80.10 &  \bf 67.23 &   \bf 81.31 &      70.63 &      61.41 &      57.44 &  \bf 29.34 &  \bf 45.87 &    374 \\
               & UDify &  \bf 88.59 &      59.22 &       72.82 &  \bf 79.37 &  \bf 67.72 &  \bf 60.13 &      21.71 &      39.25 &    374 \\
    \addlinespace
    Naija NSC & UDify &  \bf 55.44 &  \bf 51.32 &   \bf 97.03 &  \bf 45.75 &  \bf 32.16 &  \bf 31.62 &   \bf 4.73 &  \bf 29.33 &      0 \\
    \addlinespace
    North Sami Giella & UDPipe &  \bf 92.54 &  \bf 90.03 &   \bf 88.31 &  \bf 78.30 &  \bf 73.49 &  \bf 70.94 &  \bf 62.40 &  \bf 61.45 &   2.3k \\
               & UDify &      90.21 &      83.55 &       71.50 &      74.30 &      67.13 &      64.41 &      51.20 &      40.63 &   2.3k \\
    \addlinespace
    Norwegian Bokmaal & UDPipe &  \bf 98.31 &  \bf 97.14 &   \bf 98.64 &      92.39 &      90.49 &      88.18 &      84.06 &      86.53 &  15.7k \\
               & UDify &      98.18 &      96.36 &       97.33 &  \bf 93.97 &  \bf 92.18 &  \bf 90.40 &  \bf 85.02 &  \bf 87.13 &  15.7k \\
    \addlinespace
    Norwegian Nynorsk & UDPipe &  \bf 98.14 &  \bf 97.02 &   \bf 98.18 &      92.09 &      90.01 &      87.68 &      82.97 &      85.47 &  14.2k \\
               & UDify &  \bf 98.14 &      96.55 &       97.18 &  \bf 94.34 &  \bf 92.37 &  \bf 90.39 &  \bf 85.01 &  \bf 86.71 &  14.2k \\
    \addlinespace
    Norwegian NynorskLIA & UDPipe &      89.59 &      86.13 &       93.93 &      68.08 &      60.07 &      54.89 &      44.47 &      50.98 &    340 \\
               & UDify &  \bf 95.01 &  \bf 93.36 &   \bf 96.13 &  \bf 75.40 &  \bf 69.60 &  \bf 65.33 &  \bf 56.90 &  \bf 62.27 &    340 \\
    \addlinespace
    Old Church Slavonic PROIEL & UDPipe &  \bf 96.91 &  \bf 90.66 &   \bf 93.11 &  \bf 89.66 &  \bf 85.04 &  \bf 83.41 &  \bf 73.63 &  \bf 77.81 &   4.1k \\
               & UDify &      84.23 &      71.30 &       65.70 &      76.71 &      66.67 &      64.10 &      46.25 &      43.88 &   4.1k \\
    \addlinespace
    Old French SRCMF & UDPipe &  \bf 96.09 &  \bf 97.81 &  \bf 100.00 &  \bf 91.74 &  \bf 86.83 &  \bf 83.85 &  \bf 79.91 &  \bf 83.85 &  13.9k \\
               & UDify &      95.73 &      96.98 &  \bf 100.00 &  \bf 91.74 &      86.65 &      83.49 &      78.85 &      83.49 &  13.9k \\
    \addlinespace
    Persian Seraji & UDPipe &  \bf 97.75 &  \bf 97.78 &   \bf 97.44 &  \bf 90.05 &  \bf 86.66 &  \bf 83.26 &  \bf 81.23 &  \bf 80.93 &   4.8k \\
               & UDify &      96.22 &      94.73 &       92.55 &      89.59 &      85.84 &      81.98 &      76.65 &      74.74 &   4.8k \\
    \addlinespace
    Polish LFG & UDPipe &  \bf 98.80 &  \bf 95.49 &   \bf 97.54 &      96.58 &  \bf 94.76 &      93.01 &  \bf 87.04 &  \bf 90.26 &  13.8k \\
               & UDify &  \bf 98.80 &      87.71 &       94.04 &  \bf 96.67 &      94.58 &  \bf 93.03 &      76.50 &      85.15 &  13.8k \\
    \addlinespace
    Polish SZ & UDPipe &      98.34 &  \bf 93.04 &   \bf 97.16 &      93.39 &  \bf 91.24 &  \bf 89.39 &  \bf 81.06 &  \bf 85.99 &   6.1k \\
               & UDify &  \bf 98.36 &      67.11 &       93.92 &  \bf 93.67 &      89.20 &      87.31 &      48.47 &      80.24 &   6.1k \\
    \addlinespace
    Portuguese Bosque & UDPipe &      97.07 &  \bf 96.40 &   \bf 98.46 &      91.36 &  \bf 89.04 &  \bf 85.19 &  \bf 76.67 &  \bf 83.06 &   8.3k \\
               & UDify &  \bf 97.10 &      89.70 &       91.60 &  \bf 91.37 &      87.84 &      84.13 &      69.09 &      78.64 &   8.3k \\
    \addlinespace
    Portuguese GSD & UDPipe &  \bf 98.31 &  \bf 99.92 &   \bf 99.30 &      93.01 &      91.63 &      87.67 &  \bf 85.96 &      86.94 &   9.7k \\
               & UDify &      98.04 &      95.75 &       98.95 &  \bf 94.22 &  \bf 92.54 &  \bf 89.37 &      82.32 &  \bf 87.90 &   9.7k \\
    \addlinespace
    Portuguese PUD & UDify &  \bf 90.14 &  \bf 51.16 &   \bf 99.79 &  \bf 87.02 &  \bf 80.17 &  \bf 74.10 &  \bf 17.51 &  \bf 74.10 &      0 \\
    \addlinespace
    Romanian Nonstandard & UDPipe &      96.68 &  \bf 90.88 &   \bf 94.78 &      89.12 &      84.20 &      78.91 &  \bf 65.93 &  \bf 73.44 &   8.0k \\
               & UDify &  \bf 96.83 &      88.89 &       89.33 &  \bf 90.36 &  \bf 85.26 &  \bf 80.41 &      64.68 &      68.11 &   8.0k \\
    \bottomrule
    \end{tabular}
    \end{center}
    \caption{\label{table:full-results-3} The full test results of UDify on 124 treebanks (part 3 of 4).}
\end{table*}

\begin{table*}
    \fontsize{8}{10}\selectfont
    \begin{center}
    \begin{tabular}{@{}llrrrrrrrrr@{}}
    \toprule
    \sc Treebank & \sc Model & \sc UPOS & \sc UFeats & \sc Lemmas & \sc UAS & \sc LAS & \sc CLAS & \sc  MLAS & \sc BLEX & \sc Size \\
    \midrule
    Romanian RRT & UDPipe &  \bf 97.96 &  \bf 97.53 &   \bf 98.41 &      91.31 &      86.74 &      82.57 &      79.02 &  \bf 81.09 &   8.0k \\
               & UDify &      97.73 &      96.12 &       95.84 &  \bf 93.16 &  \bf 88.56 &  \bf 84.87 &  \bf 79.20 &      79.92 &   8.0k \\
    \addlinespace
    Russian GSD & UDPipe &  \bf 97.10 &  \bf 92.66 &   \bf 97.37 &      88.15 &      84.37 &      82.66 &  \bf 74.07 &  \bf 80.03 &   3.9k \\
               & UDify &      96.91 &      87.45 &       77.73 &  \bf 90.71 &  \bf 86.03 &  \bf 84.51 &      67.24 &      62.08 &   3.9k \\
    \addlinespace
    Russian PUD & UDify &  \bf 93.06 &  \bf 63.60 &   \bf 77.93 &  \bf 93.51 &  \bf 87.14 &  \bf 83.96 &  \bf 37.25 &  \bf 61.86 &      0 \\
    \addlinespace
    Russian SynTagRus & UDPipe &  \bf 99.12 &  \bf 97.57 &   \bf 98.53 &      93.80 &      92.32 &      90.85 &  \bf 87.91 &  \bf 89.17 &  48.8k \\
               & UDify &      98.97 &      96.29 &       94.47 &  \bf 94.83 &  \bf 93.13 &  \bf 91.87 &      86.91 &      85.44 &  48.8k \\
    \addlinespace
    Russian Taiga & UDPipe &      93.18 &      82.87 &       89.99 &      75.45 &      69.11 &      65.31 &      48.81 &      57.21 &    881 \\
               & UDify &  \bf 95.39 &  \bf 88.47 &   \bf 90.19 &  \bf 84.02 &  \bf 77.80 &  \bf 75.12 &  \bf 59.71 &  \bf 65.15 &    881 \\
    \addlinespace
    Sanskrit UFAL & UDify &  \bf 37.33 &  \bf 17.63 &   \bf 37.38 &  \bf 40.21 &  \bf 18.56 &  \bf 15.38 &   \bf 0.85 &   \bf 4.12 &      0 \\
    \addlinespace
    Serbian SET & UDPipe &  \bf 98.33 &  \bf 94.35 &   \bf 97.36 &      92.70 &      89.27 &      87.08 &  \bf 79.14 &      84.18 &   2.9k \\
               & UDify &      98.30 &      92.22 &       95.86 &  \bf 95.68 &  \bf 91.95 &  \bf 90.30 &      78.45 &  \bf 84.93 &   2.9k \\
    \addlinespace
    Slovak SNK & UDPipe &      96.83 &  \bf 90.82 &   \bf 96.40 &      89.82 &      86.90 &      84.81 &      74.00 &      81.37 &   8.5k \\
               & UDify &  \bf 97.46 &      89.30 &       93.80 &  \bf 95.92 &  \bf 93.87 &  \bf 92.86 &  \bf 77.33 &  \bf 85.12 &   8.5k \\
    \addlinespace
    Slovenian SSJ & UDPipe &      98.61 &  \bf 95.92 &   \bf 98.25 &      92.96 &      91.16 &      88.76 &  \bf 83.85 &  \bf 86.89 &   6.5k \\
               & UDify &  \bf 98.73 &      93.44 &       96.50 &  \bf 94.74 &  \bf 93.07 &  \bf 90.94 &      81.55 &      86.38 &   6.5k \\
    \addlinespace
    Slovenian SST & UDPipe &      93.79 &      86.28 &   \bf 95.17 &      73.51 &      67.51 &      63.46 &      52.67 &      60.32 &   2.1k \\
               & UDify &  \bf 95.40 &  \bf 89.81 &       95.15 &  \bf 80.37 &  \bf 75.03 &  \bf 71.19 &  \bf 61.32 &  \bf 67.24 &   2.1k \\
    \addlinespace
    Spanish AnCora & UDPipe &  \bf 98.91 &  \bf 98.49 &   \bf 99.17 &      92.34 &      90.26 &      86.39 &  \bf 83.97 &  \bf 85.51 &  14.3k \\
               & UDify &      98.53 &      97.89 &       98.07 &  \bf 92.99 &  \bf 90.50 &  \bf 87.26 &      83.43 &      84.85 &  14.3k \\
    \addlinespace
    Spanish GSD & UDPipe &  \bf 96.85 &  \bf 97.09 &   \bf 98.97 &      90.71 &  \bf 88.03 &  \bf 82.85 &  \bf 75.98 &  \bf 81.47 &  14.2k \\
               & UDify &      95.91 &      95.08 &       96.52 &  \bf 90.82 &      87.23 &      82.83 &      72.47 &      78.08 &  14.2k \\
    \addlinespace
    Spanish PUD & UDify &  \bf 88.98 &  \bf 54.58 &  \bf 100.00 &  \bf 90.45 &  \bf 83.08 &  \bf 77.42 &  \bf 18.06 &  \bf 77.42 &      0 \\
    \addlinespace
    Swedish LinES & UDPipe &      96.78 &  \bf 89.43 &   \bf 97.03 &      86.07 &      81.86 &      80.32 &      66.48 &      77.38 &   2.7k \\
               & UDify &  \bf 96.85 &      87.24 &       92.70 &  \bf 88.77 &  \bf 85.49 &  \bf 85.61 &  \bf 66.99 &  \bf 77.62 &   2.7k \\
    \addlinespace
    Swedish PUD & UDify &  \bf 96.36 &  \bf 80.04 &   \bf 88.81 &  \bf 89.17 &  \bf 86.10 &  \bf 85.25 &  \bf 57.12 &  \bf 72.92 &      0 \\
    \addlinespace
    Swedish Sign Language SSLC & UDPipe &  \bf 68.09 &     100.00 &  \bf 100.00 &  \bf 50.35 &  \bf 37.94 &  \bf 39.51 &  \bf 30.96 &  \bf 39.51 &     88 \\
               & UDify &      63.48 &  \bf 96.10 &  \bf 100.00 &      40.43 &      26.95 &      30.12 &      23.29 &      30.12 &     88 \\
    \addlinespace
    Swedish Talbanken & UDPipe &      97.94 &  \bf 96.86 &   \bf 98.01 &      89.63 &      86.61 &      84.45 &      79.67 &  \bf 82.26 &   4.3k \\
               & UDify &  \bf 98.11 &      95.92 &       95.50 &  \bf 91.91 &  \bf 89.03 &  \bf 87.26 &  \bf 80.72 &      81.31 &   4.3k \\
    \addlinespace
    Tagalog TRG & UDify &  \bf 60.62 &  \bf 35.62 &   \bf 73.63 &  \bf 64.04 &  \bf 40.07 &  \bf 36.84 &   \bf 0.00 &  \bf 13.16 &      0 \\
    \addlinespace
    Tamil TTB & UDPipe &      91.05 &  \bf 87.28 &   \bf 93.92 &      74.11 &      66.37 &      63.71 &  \bf 55.31 &  \bf 59.58 &    401 \\
               & UDify &  \bf 91.50 &      83.21 &       80.84 &  \bf 79.34 &  \bf 71.29 &  \bf 69.10 &      53.62 &      54.84 &    401 \\
    \addlinespace
    Telugu MTG & UDPipe &      93.07 &      99.03 &  \bf 100.00 &      91.26 &  \bf 85.02 &  \bf 81.76 &  \bf 77.75 &  \bf 81.76 &   1.1k \\
               & UDify &  \bf 93.48 &  \bf 99.31 &  \bf 100.00 &  \bf 92.23 &      83.91 &      79.92 &      76.10 &      79.92 &   1.1k \\
    \addlinespace
    Thai PUD & UDify &  \bf 56.78 &  \bf 62.48 &  \bf 100.00 &  \bf 49.05 &  \bf 26.06 &  \bf 18.42 &   \bf 3.77 &  \bf 18.42 &      0 \\
    \addlinespace
    Turkish IMST & UDPipe &  \bf 96.01 &  \bf 92.55 &   \bf 96.01 &      74.19 &  \bf 67.56 &      63.83 &  \bf 56.96 &  \bf 61.37 &   3.7k \\
               & UDify &      94.29 &      84.49 &       87.71 &  \bf 74.56 &      67.44 &  \bf 63.87 &      49.42 &      54.10 &   3.7k \\
    \addlinespace
    Turkish PUD & UDify &  \bf 77.34 &  \bf 24.59 &   \bf 84.31 &  \bf 67.68 &  \bf 46.07 &  \bf 39.95 &   \bf 2.61 &  \bf 32.50 &      0 \\
    \addlinespace
    Ukrainian IU & UDPipe &      97.59 &  \bf 92.66 &   \bf 97.23 &      88.29 &      85.25 &      81.90 &  \bf 73.81 &      79.10 &   5.3k \\
               & UDify &  \bf 97.71 &      88.63 &       94.00 &  \bf 92.83 &  \bf 90.30 &  \bf 88.15 &      72.93 &  \bf 81.04 &   5.3k \\
    \addlinespace
    Upper Sorbian UFAL & UDPipe &      62.93 &      41.10 &       68.68 &      45.58 &      34.54 &      27.18 &   \bf 3.37 &      16.65 &     24 \\
               & UDify &  \bf 84.87 &  \bf 48.84 &   \bf 72.68 &  \bf 71.55 &  \bf 62.82 &  \bf 56.04 &      16.19 &  \bf 37.89 &     24 \\
    \addlinespace
    Urdu UDTB & UDPipe &      93.66 &      81.92 &   \bf 97.40 &      87.50 &      81.62 &      75.20 &      55.02 &      73.07 &   4.0k \\
               & UDify &  \bf 94.37 &  \bf 82.80 &       96.68 &  \bf 88.43 &  \bf 82.84 &  \bf 77.00 &  \bf 56.70 &  \bf 73.97 &   4.0k \\
    \addlinespace
    Uyghur UDT & UDPipe &  \bf 89.87 &  \bf 88.30 &   \bf 95.31 &  \bf 78.46 &  \bf 67.09 &  \bf 60.85 &  \bf 47.84 &  \bf 57.08 &   1.7k \\
               & UDify &      75.88 &      70.80 &       79.70 &      65.89 &      48.80 &      38.95 &      21.75 &      31.31 &   1.7k \\
    \addlinespace
    Vietnamese VTB & UDPipe &      89.68 &  \bf 99.72 &   \bf 99.55 &      70.38 &      62.56 &      60.03 &      55.56 &      59.54 &   1.4k \\
               & UDify &  \bf 91.29 &      99.58 &       99.21 &  \bf 74.11 &  \bf 66.00 &  \bf 63.34 &  \bf 58.71 &  \bf 62.61 &   1.4k \\
    \addlinespace
    Warlpiri UFAL & UDify &  \bf 33.44 &  \bf 18.15 &   \bf 39.17 &  \bf 21.66 &   \bf 7.96 &   \bf 7.49 &   \bf 0.00 &   \bf 0.88 &      0 \\
    \addlinespace
    Yoruba YTB & UDify &  \bf 50.86 &  \bf 78.32 &   \bf 85.56 &  \bf 37.62 &  \bf 19.09 &  \bf 16.56 &   \bf 6.30 &  \bf 12.15 &      0 \\
    \bottomrule
    \end{tabular}
    \end{center}
    \caption{\label{table:full-results-4} The full test results of UDify on 124 treebanks (part 4 of 4).}
\end{table*}

\end{document}